\documentclass[
    numbers=noenddot,
    parskip=half-,
    fontsize=12pt,
    paper=a4,
    oneside,
    titlepage,
    bibliography=totoc,
    chapterprefix=false,
]{scrbook}

\usepackage{graphicx}
\usepackage[onehalfspacing]{setspace}

\usepackage{tocbasic}
\usepackage{booktabs}
\usepackage{multicol}
\usepackage{multirow}

\usepackage[]{scrlayer-scrpage}

\usepackage[citestyle=alphabetic, bibstyle=alphabetic, sorting=nyt, backend=biber, language=english, backref=true, maxcitenames=2]{biblatex}

\usepackage[autostyle,english=american,german=quotes]{csquotes}
\usepackage[titletoc]{appendix}

\graphicspath{{images/}}

\usepackage{float}

\title{TITLE}

\author{AUTHOR}

\date{\today}

\newcommand{\thesisType}{Masterarbeit}

\makeatletter
\let\thetitle\@title
\let\theauthor\@author
\let\thedate\@date
\makeatother

\begin{document}

\frontmatter
\begin{titlepage}
    \centering
    \begin{onehalfspace}
    	
        	\includegraphics[width=7cm]{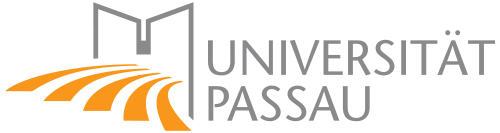}\\
        	\vspace{1.0cm}
        	\large {\bfseries Lehrstuhl f\"ur Data Science }\\

        	\vspace{2.5cm}

            \begin{doublespace}
            	{\textsf{\Huge{Neural Models for Offensive Language Detection}}}
            \end{doublespace}

        	\vspace{2cm}

            \Large{Masterarbeit von}\\

        	\vspace{1cm}

        	{\bfseries \large{Ehab Hamdy}}

        	\vfill

        	{\large
        		\begin{tabular}[l]{cc}
        			\textsc{1.~Pr\"ufer} & \textsc{2.~Pr\"ufer} \\
        			Prof.~Dr.~Jelena Mitrović& Prof.~Dr.~Michael Granitzer
        		\end{tabular}
        	}

        	\vspace{1.5cm}

        	\parbox{\linewidth}{\hrule\strut}

            \vfill

	    \thedate
    \end{onehalfspace}
\end{titlepage}

\tableofcontents
\newpage

\chapter*{Abstract}

Offensive language detection is an ever-growing natural language processing (NLP) application. This growth is mainly because of the widespread usage of social networks, which becomes a mainstream channel for people to communicate, work, and enjoy entertainment content. Many incidents of sharing aggressive and offensive content negatively impacted society to a great extend. We believe contributing to improving and comparing different machine learning models to fight such harmful contents is an important and challenging goal for this thesis. We targeted the problem of offensive language detection for building efficient automated models for offensive language detection. With the recent advancements of NLP models, specifically, the Transformer model, which tackled many shortcomings of the standard seq-to-seq techniques. The BERT model has shown state-of-the-art results on many NLP tasks. Although the literature still exploring the reasons for the BERT achievements in the NLP field. Other efficient variants have been developed to improve upon the standard BERT, such as RoBERTa and ALBERT. Moreover, due to the multilingual nature of text on social media that could affect the model decision on a given tween, it is becoming essential to examine multilingual models such as XLM-RoBERTa trained on 100 languages and how did it compare to unilingual models. The RoBERTa based model proved to be the most capable model and achieved the highest F1 score for the tasks. Another critical aspect of a well-rounded offensive language detection system is the speed at which a model can be trained and make inferences. In that respect, we have considered the model run-time and fine-tuned the very efficient implementation of FastText called BlazingText that achieved good results, which is much faster than BERT-based models.
\newpage

\chapter*{Acknowledgments}

I want to express my most profound appreciation to my supervisors. I immensely appreciate and thank Prof. Dr. Jelena Mitrovic for her guidance throughout my master's degree and thesis writing. I have received incredible support and very kind and constructive cooperation that always pushed me to work harder while maintaining an extremely positive mindset and hope for a better future. It is a genuine pleasure to express my most tremendous thanks to Pro. Dr. Michael Granitzer for the invaluable knowledge, insights that helped me while writing this dissertation. I am genuinely thankful for his supervision and constructive feedback to make my work stand out.
 
I am deeply thankful to all the lecturers and staff members at Passau University for their efforts to provide top-notch technical suggestions while pursuing my master's degree.
 
I also owe a specially thanks to my family and friends for their constant encouragement.
\newpage

\thispagestyle{empty}
\cleardoublepage
\listoffigures
\newpage

\thispagestyle{empty}
\cleardoublepage
\listoftables
\newpage

\mainmatter

\chapter{Introduction}\label{chap:introduction}

In the era of social media and communications, it is easier than ever to freely express opinions on a plethora of topics. This openness creates a proliferation of useful information for productivity and making the right decisions. According to a statista report, the number of active social media users worldwide have just exceeded four billion user, that is more than half of the world population. The user base is also expected to grow steadily during the next five years. Studies shows that vulnerable groups of children and adolescents represents a large portion of social media users.

With the greatness of social media benefits, unfortunately it also brought up opportunities for harsh discussions that can easily reach uncivilized, hateful, offensive or toxic levels \cite{shaw2011hate}. The pervasive use of hate inducing speech and varies offensive expressions on the internet has led to poisoning effects such as reduced productivity due to users abstaining from sharing useful opinions, users depressions, and in sever cases led to hate crimes with devastating effects on the community. Fighting back those negative effects requires a collaborative efforts which starts with limiting the spread of offensive language and even proactively monitor highly suspicious users on social networks.

\section{Motivation}
With the rapid adoption of social media platforms between billions of people around the world, an astronomical scale of user generated contents has been created. Unfortunately these platforms can easily used for online abuse and harassment that can leave serious effects on its victims. Therefore the offensive content detection becomes a great concern for online communities and social media platforms. According to analysis conducted at \cite{hinduja2010bullying}, youth who experienced cyberbullying as either victims and offenders were twice as likely to attempt suicide than those who had no experienced such form of peer aggression. Mitigating such negative effects requires finding solutions to early detect and monitor hate speech on the cyberspace before any major escalations and propagation of negative thoughts beyond the internet walls.

Given the enormous scale of internet users and the gigantic inflow of online contents, manual detection of hate speech is deemed ineffective. Therefore the necessity of auto detection and removal of offensive and profane language in online environments. Advancements in machine learning and natural language processing, specifically speaking \textit{transformer based models}
showed outstanding results in all kinds of text processing including text translation, natural language modeling, sentiment analysis. Offensive language detection is considered a sentiment analysis task, where an offensive language is a statement with negative sentiment and normal language can be considered a positive sentiment. Offensive language detection tasks can also include categorization of text sentiment into further classes to detect the types of offense, the targets and many other aspects of the phenomena. 

\section{Granularity and sensitivity of Violating Content}
In this section we are using the term \textit{Violating Content} as a wide term for any type of form of contents with potential harm to the online community. The term \textit{Violating Content} and \textit{Not Allowed Content} are the terms used by popular social network platforms such as Facebook and Twitter. These terms are used to allow social networks to set its varying standards for harmful contents on the internet. Although profanity words is considered a major form of Offensive Language, major platform doesn't classify it as a harmful content but as freedom of speech unless it implies physical harm to individual or group of people and associated with other types of violations such as suspected association with terrorist groups or other violating media (images, audio or videos). With the broad definition of not allowed content versus violating content, social network greatly relies on explicit reports from active users to take down violating contents. Other communities such as online learning forums or educational institutions online portals have stricter measure of offensive language that would not allow permitted contents of general social media platforms. Consequently, there is no one ideal model that would fit every platform on the internet. The common distinction levels are as following:
\begin{enumerate}
\item \textbf{Offensive Versus Non-offensive}: This is the most stricter measure when offensive speech can include all forms of potential harmful. This is the mostly desired by communities of specialized purposes such as online education platforms, while is less preferred and even considered against freedom of speech from the point of view of general purpose communities such as Facebook, Twitter and others.
\item \textbf{Offensive Versus Hateful speech Versus Normal language}: This form of detection provide a distinction between usage of profanity in particular context against other forms of hate speech such as harassment or attacks against race, ethnicity, religion, gender, or disability.
\end{enumerate}

Offensive language as the general categorization of all kinds of potentially harmful contents, explicit or implicit can also be classified in different levels. According to our review of recent literature, offensive language can be classified for either type identification, target identification of offense.
\begin{enumerate}
\item Based on the \textit{Type identification}, offensive language is classified into; targeted insult where a statement contains insults or threat to an individual, a group, or others. Or untargeted insults is a statement with general and non targeting profanity or swearing.
\item While Target Identification, is the task of distinguishing the target of insult, an insult could target an individual such as a famous person, a named individual or an unnamed person interacting in the conversation. Also an insult that target a group of people considered as a unity due to the same ethnicity, gender or sexual orientation, political affiliation, religious belief, or something else. Or an insult the target other entities such as events, issues, or situations.
\end{enumerate}

Violating contents are not only associated with the choice of words but also depend on the context where these words appeared in. For example using gender exclusive language to control membership in a health or positive support group such as breastfeeding groups for women only is allowed given the users shows a clear indication of their intents. These considerations make the problem of offensive language detection even more interesting and challenging.


\section{Neural Networks approaches for the offensive language problem}
In the recent years, the advancements in neural network approaches have witnessed a big leap forward in terms of language understanding. Starting from basic recurrent neural networks to advanced transformers model that have harnessed the parallalization capabilities of modern hardware and allowed for transfer learning approaches via pretraining general models that can be fine-tuned for varies other tasks. These ultimately contributed to state of the art results and shorter development cycles when developing models for specific tasks. In this thesis we are following a data science workflow to tackle the problem of offensive language detection.

The term data science refer to the complete development life cycle of machine learning model starting with data collection and cleaning, data preprocessing, incorporation of statistical and machine learning methodologies, and programming approach to train and deploy highly accurate models that can be integrated into ubiquitous systems with the goal to improve its efficiency or mitigating existing issues.

\begin{figure}[H]
	\centering
	\includegraphics[width=12cm]{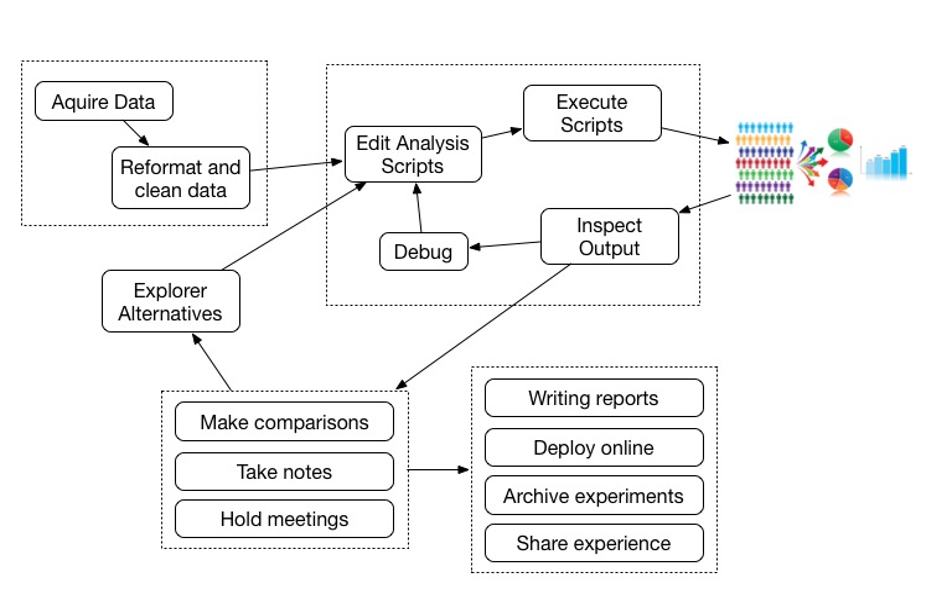}
	\caption{Data science workflow}
	\label{fig:dsworkflow}
\end{figure}

A data science workflow where we start by looking at the data and then determining the obvious issues and relevant data points. Following that an inspection stage for listing the most anticipated issues in the data it is very crucial to make the data usable using cleaning and fixing methods. Afterwards, the stage to extract summaries and preliminary reports becomes very important so that we can inspect the questions that can be answered from the extracted description of the data. Next is the stage of thoroughly clean and filter the datasets and loading it to the adequate tools and frameworks. If problems where detected while loading the data we step back to the cleaning stage, otherwise we start to explore it using common descriptive statistical analysis techniques. The cycle continues to create machine leaning models, assessing and testing performance measures. In parallel to creating models we create visualizations that help telling stories about data and communicate and discuss the outcomes. Finally, the results are deployed and shared with the community. Figure \ref{fig:dsworkflow} depicts the stages of data science work flow.

\section{Contribution}
To address the challenge of offensive language detection, we propose transfer learning based approaches, specifically Bidirectional Encoder Representations from Transformers (BERT) which showed top results in the OffensEval challenge for offensive language detection based on text (tweets) from tweeter platform. Secondly, we will show the utilizing of incredibly fast model called BlazingText that can be trained on millions of text tweets and achieve respectable results while only taking few seconds for the training process and also blazing fast inferences that are unmatched by neural network based models. 

\section{Thesis Structure}
In this thesis, we will introduce a thorough literature review of the existing studies that tackled the problem of offensive language detection. In chapter 3, we are going to discuss the models in details we have used and how BERT based models are enhanced and fine-tuned for the offensive language detection task, in addition to that we will describe our contribution to the offensEval 2020 challenge and how the AWS BlazingText is utilized to harness the power of the cloud. In Chapter 4, we will present the data set used in our experiments and a comparative analysis of the results. In chapter 5, will have a discussion on the results, limitation and suggestions for future research and finally a conclusion with a summary of contribution.

\chapter{Background}\label{chap:background}
Identifying offensive language on the internet has become one of the most popular applications of NLP \cite{schmidt2017survey}. The authors in \cite{baldwin2013noisy}  conducted a systematic study on texts of social media and microblogs. They proved that text on social media texts is linguistically noisy and less grammatical than edited text. Researchers have proposed a set of feature engineering methods with different levels of effectiveness. Surface features such as bag of words and n-grams are the simplest representative features of texts. A more advanced technique for extracting numerical representations for texts is word generalization which include methods such as word clustering \cite{warner2012detecting}, distributed word representations using neural networks (word embedding), and document embedding \cite{djuric2015hate}. Another prominent feature engineering technique is multimodal features, which incorporates additional features from images, audio and video. Authors in \cite{yang2019exploring} used feature fusion of text and photos for detecting hate speech in Facebook posts.

Furthermore, the advancements of language modeling and machine learning techniques show promising results when tackling the problem of offensive language identification. In the  first edition of OffensEval \cite{zampieri2019semeval}, transfer learning methods such as BERT \cite{devlin2019bert} proved to be among the most effective and accurate (as shown by the best system at OffensEval 2019~\cite{liu-etal-2019-nuli}) especially when dealing with limited labeled datasets. Still, some other models proved their predictive power as well, e.g., the C-BiGRU model which combines a Convolutional Neural Network (CNN) with a bidirectional Recurrent Neural Network (RNN) \cite{mitrovic2019nlpup} that scored in the 9th position of OffensEval 2019.

Training and inference time is also an important aspect of a machine learning model. In often cases there is a trade of between how accurate a model is and how fast it can train and make predictions while it is deployed. Popular techniques were invented to tackle machine learning problems very fast while maintaining a respectable prediction accuracy and few shortcomings. Popular models such as FastText \cite{bojanowski2017enriching} and BlazingText \cite{gupta2017blazingtext} are specifically designed for speed and enable training a model on very large datasets.

\section{Data Preprocessing}

One of the fundamental building blocks of many NLP tasks and its applications is text preparation and preprocessing. That is to accommodate the typical nature of social media short texts of being informal \cite{kontopoulos2013ontology}, containing irregularities and the excessively use slang and colloquial language. Data preprocessing can also contribute to disambiguation and removing non-indicative parts of sentences of short tweets , thus, improving the accuracy of machine learning models operating on a preprocessed data sets compared to raw texts. Table \ref{table:tweetsexamples} shows an example of tweets from the OffensEval dataset that represents a typical social media text format.

\begin{table}[h]
\centering 
\begin{tabular}{|l|l|}
\hline
Tweet                                                                                                                                                                                                                                                                                                       & Classification \\ \hline
\begin{tabular}[c]{@{}l@{}}\#ConstitutionDay is revered by Conservatives, hated by \\ Progressives/Socialist/Democrats that want to change it.\end{tabular}                                                                                                                                                 & Normal         \\ \hline
\begin{tabular}[c]{@{}l@{}}\#BiggBossTamil janani won the task. She is going to first \\ final list :clap::clap::clap:\end{tabular}                                                                                                                                                                                    & Normal         \\ \hline
\#StopKavanaugh he is liar like the rest of the \#GOP URL                                                                                                                                                                                                                                                   & Offensive      \\ \hline
\begin{tabular}[c]{@{}l@{}}\#Benghazi  \#Haiti  \#UraniumOne  \#SethRich What more\\ does the justice department need? No double standards! \\ Just because you're Rich and Powerful doesn't mean you \\ can rob, seal and murder! \#DrainTheSwamp \\ \#HillaryForPrison  \#VoteRed \#MAGA URL\end{tabular} & Offensive      \\ \hline
\end{tabular}
\caption{\label{table:tweetsexamples} Example of Tweets from the OffensEval dataset}
\end{table}

\begin{figure}[h]
	\centering
	\includegraphics[width=14cm]{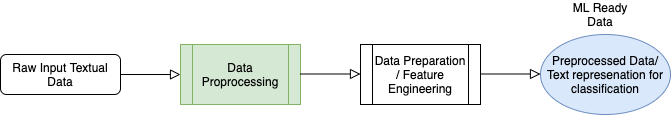}
	\caption{High level overview of the processing pipeline from raw to a machine learning ready data}
	\label{fig:procpip}
\end{figure}

After the preprocessing stage, it is necessary to transform the textual information into a numerical representation that preserve the semantic relationships between words in order to feed it to a machine learning algorithm. This process is called feature engineering which we will discuss in the next section. Figure \ref{fig:procpip} shows a high level overview of the processing pipeline needed for bringing text data into a compatible version for machine learning algorithms.

Text extracted from social media contains a wide range of irregularities such as misspelling, duplication and incorrect punctuation; also abbreviations, URLs and genre specific words which are commonly seen in texts with offensive language connotation. Text preprocessing operations are completed with two main targets or levels:
\begin{enumerate}
  \item \textbf{Sentence structure level:} processing text targeting its structures involves aspects such as punctuation - where decision such as dealing with duplicate punctuation or whether they should be completely removed. Another aspect is the consideration of exclusive grammatical structures while ignoring other structures.
  \item \textbf{Vocabulary level:} working on the vocabulary level is considered to be more critical due to its greater influence of the final results of NLP tasks. This involves decisions such as case normalization which have direct influence on the vocabulary size and efficiency of a machine learning model.
\end{enumerate}

The following list include common text normalization processes that are useful for wide range of NLP application:

\begin{itemize}

  \item \textbf{Tokenization:} is a key technique when working with text data where a sentence is separated into smaller units such as characters, words, or subwords. 
  \item \textbf{Case conversion:} one of the common text preprocessing steps is to unify all text cases such that all characters be lower or upper case, while the most commonly conversion is to lower case letters. That way repeated words with different cases would come down to only one word thereby reducing vocabulary size and making a machine learning model approach out objective much better.
  
  \item \textbf{Word normalization:} this process involves processes such as stemming, lemmatization, fixing misspelling and converting rare words or spelling variants into commonly used vocabulary. A common approach to handle word normalization is by utilizing a dictionary to map words for conversion.
  \item \textbf{Punctuation and special character removal:} non alphanumeric characters typically don't add a value to the text understanding process and introduce noise that negatively affects a machine learning model performance. Therefore, cleaning up text data from these characters usually hold great benefits for the model accuracy and speed.
  \item \textbf{Special Symbols transformation:} this include replacing Emojis to a representative text which can greatly enhance the context and change the meaning of the sentence or assert the intention of the writer. Studying the correlation between the sentiment of a sentence and special symbols such as emojis is also important.
  \item \textbf{Segmentation:} social media text are famous for hashtags which is a prominent example where we need segmentation by separating concatenated hashtags into individual words, thus enriching the meaning of the sentence.
  \item \textbf{User mentions and links normalization:} this include removing duplicate user mention and links. Another common link processing is replacing link with the string "HTTP" since it has a word embedding in most pretrained models.
\end{itemize}




\section{Feature Extraction Methods}
After obtaining a cleaned and normalized version of the textual dataset, the next step in an NLP application pipeline is to extract appropriate representation of the text in the form of \textit{numerical format} that is suitable for predictive machine learning models. The main goal of extracting high quality features from the text is to maintain the contextual and semantic relationships as close as possible. A representation vector of a document can be an average of the vector representation of the words in that document, or it can be a learned representation for the whole document/sentence which can be done with a representation technique such as Doc2Vec. We will discus the different feature representation techniques in more details later in this section. 

The following shows an example representation, commonly called word embedding of the word "king" from a GloVe model. Looking at the victor number doesn't necessary have a particular meaning by itself but it holds the correct distance similarity to the vector representations of other words in the language. 

[ 0.50451 , 0.68607 , -0.59517 , -0.022801, 0.60046 , -0.13498 , -0.08813 , 0.47377 , -0.61798 , -0.31012 , -0.076666, 1.493 , -0.034189, -0.98173 , 0.68229 , 0.81722 , -0.51874 , -0.31503 , -0.55809 , 0.66421 , 0.1961 , -0.13495 , -0.11476 , -0.30344 , 0.41177 , -2.223 , -1.0756 , -1.0783 , -0.34354 , 0.33505 , 1.9927 , -0.04234 , -0.64319 , 0.71125 , 0.49159 , 0.16754 , 0.34344 , -0.25663 , -0.8523 , 0.1661 , 0.40102 , 1.1685 , -1.0137 , -0.21585 , -0.15155 , 0.78321 , -0.91241 , -1.6106 , -0.64426 , -0.51042 ]

Learning high quality word embedding requires training a model on large amounts of text corpus such as Wikipedia articles \footnote{Here is a list of popular large text corpora freely available on the internet https://en.wikipedia.org/wiki/List\_of\_text\_corpora}. This training process can very time consuming. However, since training a word representation model is a one time task, there are many researchers who provided pretrained model for word representation that ensure high quality representation for many machine learning task. There are some observations regarding the direct usage of these representation on specific domains and in particular to represent text in social media, since the existing pretrained models are trained based on formal text unlike the social media colloquial language. In the following subsections we will present the popular methods of feature extraction that was commonly used in previous offensive language detection work and how some of these methods would deal with the specifications of social media text via in-domain model fine tuning.

\subsection{Bag Of Words (BOW)}
BOW is a text representation techniques in which the output is a vector representing a document in the corpus. Features are represented as a vector of length is number of unique words in the corpus. BOW has different variations including one-hot-encoding and TFIDF which will be discussed in the following sections.

\subsection{One Hot Encoding}
\begin{figure}[h]
	\centering
	\includegraphics[width=8cm]{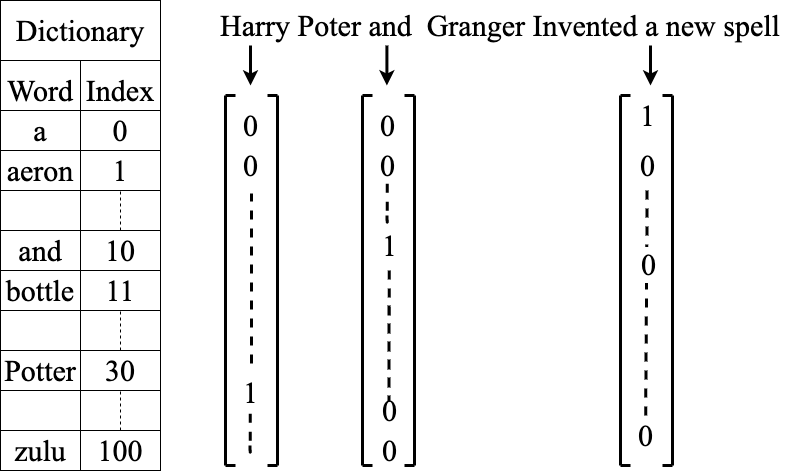}
	\caption{An example of one hot encoding using a dictionary of 100 words}
	\label{fig:eneencoding}
\end{figure}

Considered one of the basic ways to represent categorical data such as text into numerical vectors. One hot encoding involves building an indexed dictionary of words from the text corpora and representing each word with a vector of zeros with the same length as the created vocabulary dictionary while replacing the index of the word with one. Illustration \ref{fig:eneencoding} shows an example of how one hot encoding works. There are two major drawbacks when using this representation techniques for classification tasks, first, the distance between any two words representation are exactly the same which means losing context and meanings of the words. while the second is high cardinality and suffering from  curse of dimensionality. While one hot encoding shows unsuitability with classification tasks, it is effectively used to representing categorical data labels, also for its great simplicity it is used for educational purposes to get basic intuition of the concepts of text representations. One hot encoding is also a foundation for other efficient representation techniques.

\subsection{TFIDF term frequency–inverse document frequency}
TFIDF is a statistical method that is particularly concerned to reflect how important a word to document given a corpus based on the word frequency relative to the number of documents to account for the fact that some words appear more frequently in general. Calculating the TFID is done in two parts, first we calculate the term frequency for the document and then down weight it with the inverse term document frequency of the term which is calculated as:

\begin{equation} \label{eqn}
log (\frac{\# Documents}{1 + \# Documents using word}) 
\end{equation}

That way very frequent words such as “a”, “the”, “is” in English which they carry very little meaning of the actual document content won't overshadow important words when it is feed to a classifier. The output feature vector length is the number of unique words in the corpus where the value of the corresponding word is the TF-IDF calculation or zero if it didn't exist in the corresponding document. Vector length could be limited to words with a maximum threshold if needed for some applications to reduce computational complexity from curse of dimensionality.

\subsection{Word2Vec}
Word2Vec or W2V \cite{mikolov2013efficient} is a neural network based model for learning high quality continuous vector representation from a large corpus of text consisting of billions of words. Word representation from a W2V model preserves the semantic similarity between words. While BOW and TFIDF designed to capture a representation of a document in the corpora, W2V defines a representation for each word that reflect its meaning and relationships to other words. W2V was successfully used to mitigate the \emph{curse of dimentionality} problem which was present in one hot encoding. It utilizes either a Continuous Bag of Words architecture or a Skip-gram neural network architecture to learn the underlying representations for each word. Both architectures are based on a fake task where each pair of words in the corpora are taken to teach the model that they are co-occurrences. The words are first represented using One Hot Encoding to be feed into the W2V nerual network and the trained weights are considered the final representation where each word representation is a vector of the same length as the hidden layer of the W2V neural network.


\subsection{Doc2Vec}
In contrast to Word2Vec, Doc2Vec \cite{le2014distributed} is a neural network based model to learn representation of an entire document in the corpora instead of individual words. This is especially important for offensive language detection since our goal is to classify a whole sentence. While having word representations of a given document from a W2D enables us to apply techniques such as averaging word vectors of a give document to get a vector representation to a document, Doc2Vec shows a higher quality representations when depecting the relationships between the documents and it outperformed W2D averaging in several text classification and sentiment analysis. Gensime \footnote{Gensim library for text representation using W2V and D2V https://radimrehurek.com/gensim/} provide a popular implementation of the W2V and D2V .

\subsection{Global Vectors for Word Representation (GloVe)}
GloVe \cite{pennington2014glove} is an unsupervised learning model for extracting vector representations of words. Gloves uses matrix factorization technique on the word-context matrix for representation which shows better performance on analogy tasks than Word2Vec since it doesn't utilize the statistics of the corpus as they train on separate local context windows instead of the global co-occurrence counts. 

\subsection{Embeddings from Language Models (ELMo)}
ELMo is a context based model for learning word embeddings. In contrast to word embeddings from Word2Vec where it is possible to download a list of words and there fixed embedding representation that captures the meaning, syntactic and grammatical relationships between them, ELMo aims to induce word representation based on the position of the word in a sentence thus capturing the meaning and context in the representation. Using a bi-directional LSTM the model is able to create dynamic embeddings for a given word. ELMo is a LSTM based architecture trained on large dataset to predict the next word in a sequence of words in an unsupervised manner.


\subsection{ULM-FiT}
Inspired by the success of transfer learning capabilities of ImageNet models \cite{sharif2014cnn}, ULMFiT \cite{howard2018universal} offers pretrained Language Model and fine-tuning for downstream classification task. Thus creating the ground for further innovations in creating state of the art pretrained models that can be fine-tuned for tasks such as BERT and other transformer based models.


\subsection{Transformers and BERT}

\textbf{BERT} is a technique for Natural language processing developer by Google \cite{devlin2018bert}. The authors describes the BERT model as 

BERT stands for Bidirectional Encoder Representations from Transformers. It is designed to pre-train deep bidirectional representations from unlabeled text by jointly conditioning on both left and right context. As a result, the pre-trained BERT model can be fine-tuned with just one additional output layer to create state-of-the-art models for a wide range of NLP tasks.

Bert is based on the encoder of a transformer network which consists of a series of self-attention layers, there are two architecture, a small BERT that uses \textbf{12 main layer} referred to as \textbf{BERT-BASE} and the larger architecture which incorporate \textbf{24 main layers} referred to as \textbf{BERT-LARGE}. Both types are trained on corpus of 800M from BOOKSCORPUS and another version trained on 2,500M words from Wikipedia-English corpus.
        
BERT key contributions are based on two tasks:
\begin{enumerate}
\item \textbf{Masked Language Model (MLM)}: For every sentence in the corpus, 15\% of the words in a sentence is masked by replacing it with a special MASK token and the objective of the network is to predict these masked words. 

\item \textbf{Next Sentence Prediction (NSP)}: In this task, two sentences are concatenated with a SEP token and fed to the network with the objective to predict whether the first sentence is a candidate preceding sentence to the second sentence. 
\end{enumerate}

MLM and SNP are the "Fake Tasks" or the Pretraining Tasks. Bert is trained to perform these tasks mainly to enforce a rigorous understanding of the language. After the training of these two tasks are accomplished during the pre-training phase, the portion of the model for these tasks can be discarded and replaced with layers that is usefull for tasks such as classification, named entity recognition or question answering.

The main catching points of MLM and SNP tasks are that it can be trained without any labeled data, such that any raw text from Wikipedia or similar resources can simply be used for training the task. Also the tasks are challenging which enforce the BERT model to gain strong language understanding to perform well on them.

Beyond the main layers of the BERT model (12 for the BERT-base or 24 for Bert-large), Each of MLM and NSP add a single classification layer to the output of BERT to perform their corresponding tasks. As mentioned, the these final classifier layers for MLM and NSP can be replaced with task specific model that leverage the power of BERT.



\begin{figure}[H]
	\centering
	\includegraphics[width=12cm]{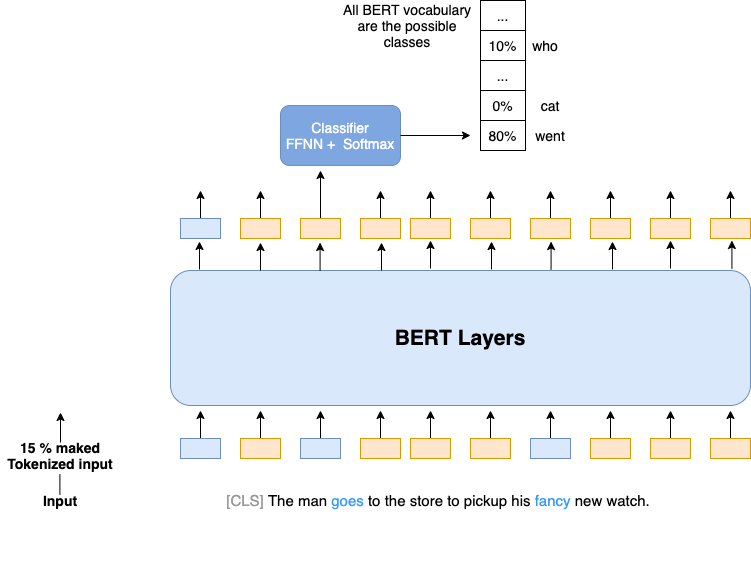}
	\caption{Visualization of the BERT Masked Model Predictions}
	\label{fig:nextSentence}
\end{figure}

\begin{figure}[H]
	\centering
	\includegraphics[width=12cm]{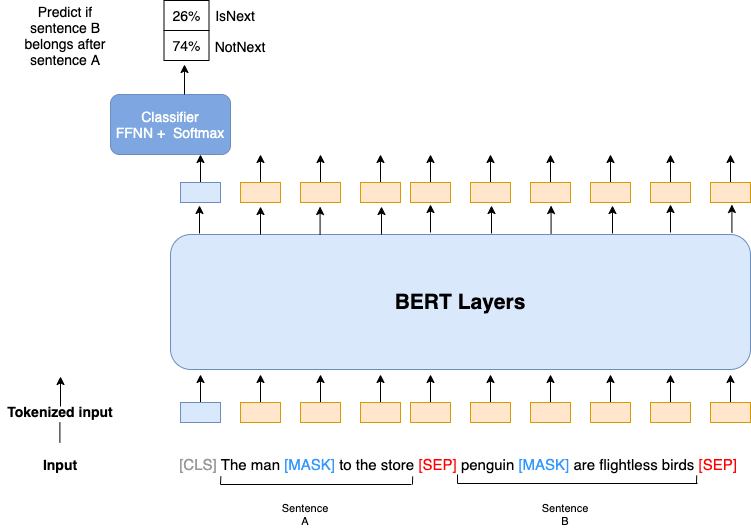}
	\caption{Visualization of Next Sentence Predictions Task}
	\label{fig:maskedModel}
\end{figure}

\section{Data Augmentation}
Data augmentation is a technique widely used to generate more data in the computer vision field. The techniques involves generating variations of the image by applying custom transformations such as cropping, rotating, adding noise or other varies methods. Data augmentation can give the model a performance boost and therefore it is also utilized in varies NLP tasks. However, data augmentation in NLP is a challenging tasks compared computer vision field since that not every word has a direct synonym and changing a word in a sentence could potentially change the context and the meaning of the sentence.  

The main goals of data augmentation are:
\begin{itemize}
\item Increase the data set size, neural network models performance largely depends on the size of the dataset.
\item Resolving class imbalance/skewness in the dataset, when training on a dataset with a largely different distribution of classes could lead to learning bias problem where the model learn to predict the outcome for a class compared to other due to its overrepresentation in the training set. In this case generating more example for the underrepresented class. Class skewedness is highly present in most dataset used for offensive language detection.

\end{itemize}
There are varies techniques for data augmentation in NLP, here are some popular methods that is also usually used in the problem of offensive language detection:
\begin{itemize}
\item \textbf{Back Translation:}  Many researchers have used the technique of back translation to generate more examples for the offensive. By translating the existing examples of offensive tweets to from one language to another and then back to the original dataset language or by translating examples from other datasets of other language, the underrepresented class can be balanced for better results. The per. Backtranslation usually yields the same meaning sentence with alternative words and sometime different sentence structure. The authors at \cite{ibrahim2020alexu} have applied back translation to balance the classes in the OLID dataset using Google Translation API, however they encountered some issues with the sentences generated from translating from the Arabic language since the back translate words was not offensive and thus effectively changed the sentence class however the technique succeeded when back-translating from many other languages such as Spanish, German, Portuguese and Italian. This is because the quality of translation between languages are different for a given model. The author reported best model F1 score when using backtranslation from Portuguese. 
\item \textbf{Word Substitutions:} By substituting the words to create a novel sentence with the same meaning of the original sentence. There are varies strategies to perform work substitutions including; Thesaurus based substitutions, word embedding substitutions, marked language substitutions and TF-IDF substitutions. 
\item \textbf{Text Generation:} The recent advances of generative models such as the Generative Adversarial Networks makes the text generation for data augmentation an attractive technique.

\end{itemize}

\section{A review of Offensive Language Detection literature}
Offensive language detection has been under the radar of many NLP researchers during the last decade. The advancements in NLP techniques with common tasks were very promising to deal with the challenges of detecting hatespeech in social media. SemEval \cite{zampieri2019semeval} \cite{zampieri-etal-2020-semeval} have attracted many researchers for the OffensEval shared task to encourage continuous work in offensive language detection. In the next subsections, we will describe the tasks of offensEval, the associated dataset, and present a review of the methods that achieved the best results.

\subsection{OffensEval Tasks Description}
OffensEval 2020 provides a massive dataset compared to the OffensEval 2019 round of the challenge which only contains 13420 records. There are many opportunities now for the state of the art deep learning models which tend to perform better on large amounts of data. At the same time this generated a new challenge on how to process the data as efficiently and as fast as possible. It becomes a necessity to rely on a GPU powered machine to train complex neural network models with billions of words in the given text corpus. In the following sections of this paper, we will illustrate our methodology to tackle that challenge and harness the usefulness of data proliferation.

\begin{figure}
	\centering
	\includegraphics[width=9cm]{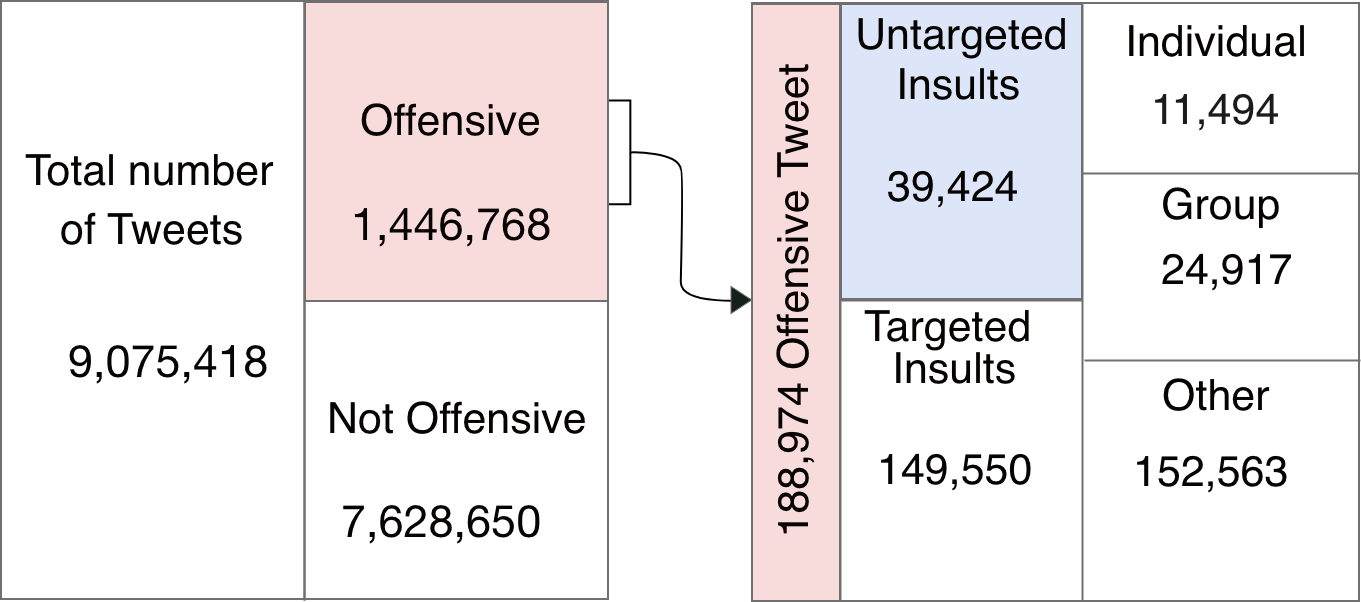}
	\caption{OLID 2020 Data and label organization based on an 0.5 threshold for label assignment}
	\label{fig:data20}
\end{figure}

The training and development data set for OffensEval 2020 is composed of over 9 million records, each record containing a short text representing a tweet. Similarly to OffensEval 2019 dataset\cite{zampieri2019predicting}, OffensEval 2020 \cite{rosenthal2020} is annotated with a hierarchical  three-level annotation schema. The goal of the first annotation level is to discriminate between offensive and non-offensive tweets. The second annotation level is based on the of offense seen as a targeted insult (TIN) or un-targeted insult (UNT). Finally, the third annotation level is focusing on three target types including individual (IND), group (GRP) and other (OTH). A summary showing the statistics of the dataset is shown in Figure \ref{fig:data20}. As the dataset labels where given in terms of probabilities, in the next section we will discuss how the label assignment was performed.

\subsection{Methodology reviews on OffensEval shared task}
In this section we will discuss the most prominent paper submitted during the 2019 and 2020 rounds of the OffensEval shared task. This review will also cover other previous work for offensive language detection and be used to further lay the ground for the methodologies we will be using in our work:

In \cite{tawalbeh2020keis}, the KEIS@ JUST team participated in the 2020 OffensEval shared task, they used a weighted ensemble consists of Bi-GRU and CNN followed by Gaussian noise and global pooling layer multiplied by weights to improve the overall performance for the Arabic and English datasets. While they also used a transfer learning from BERT beside the recurrent neural networks such as Bi-LSTM and Bi-GRU followed by a global average pooling layer for the Turkish, Danish and Greek dataset. Word embedding and contextual embedding have been used as features. Data augmentation has been used only for the Arabic language and relied on the AraVec embedding \cite{soliman2017aravec}. The best results for the KEIS@JUST team ranked 11th place out of 56 teams with 86.55\% F1-macro in the Arabic language, ranked 12th place out of 39 teams with 76.1\% F1-macro in the Danish language, ranked 28th place out of 37 teams with 76.1\% F1-macro in the Greek language, ranked 32th place out of 46 teams with 73.3\% F1-macro in the Greek language. The authors at \cite{hussein2020nlp_passau} and \cite{birkeneder2018upinf} also utilized model based on a combination of a convolutionl neural network and a recurrent neural network. At \cite{hussein2020nlp_passau} the authors used C-BiGRU which is a convolutional neural network along side a GRU based based recurrent neural network during the OffensEval 2020 competition. The experiments by the authors shows a significant performance gain of when using FastText word embeddings based on Skipgram mechanism over Word2vec embeddings, the author emphasized the superiority of the FastText to obtain accurate embeddings for rare words which was a major factor of the performance improvements.

\begin{figure}[h]
	\centering

	\includegraphics[width=15cm]{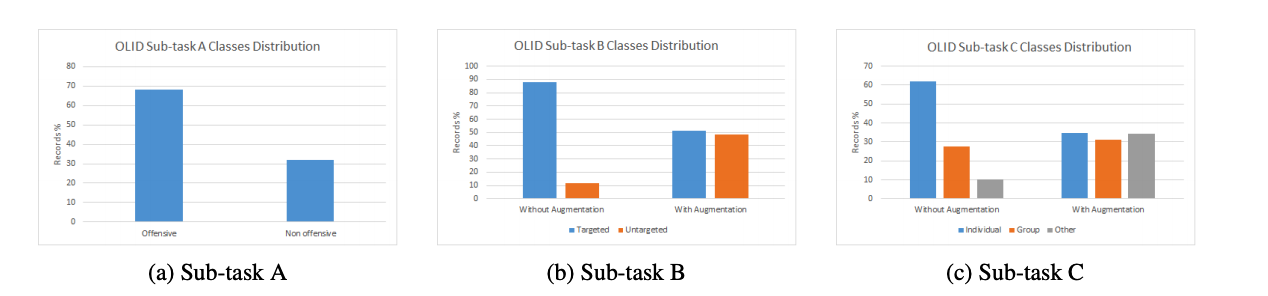}
	\caption{Classes Distribution in the Three Levels (Sub-tasks) of OLID Dataset}
	\label{fig:classdist}
\end{figure}

The work at \cite{ibrahim2020alexu} focused on solving the problem of imbalanced classes using backtranslation based augmentation introduced in \cite{sennrich2015improving}. The technique involves generating new data instances for the class that is underrepresented in the dataset by translating sentence to other languages and then translating back to English which held a different sentence with the overall same meaning and then assigning the same class. That method significantly improve the model performance. They have used Google translate API to perform the processing. The authors applied data augmentation for task B and task C dataset since they suffer more from imbalanced classes problem. Figure \ref{fig:classdist} shows a comparison of the resulting class distribution after back translation. For the classifying offensive language the authors utilized a fine tuned BERT in an ensemble of deep learning classifier achieving a 91.393\% F1 score for the Task A.

The authors at \cite{sotudeh2020guir} used BERT based model in OffensEval 2020 for offensive language detection. Since BERT is pretrained on formal text (wikipedia and books) they started with fine tunning the model to our domain. This is accomplished by taking the original model and continuing to train the masked language model and next sentence prediction objectives using a large amount of unlabeled tweets. For predictions they have using an ensemble  using the multi-view SVM with the addition of features from the contextualized language model classifier. Specifically, linear SVM classifier using various n-gram ranges are first trained for each task in addition to the BERT-based classifier. Then the output of the view-classifiers are concatenated as a final linear SVM classifier.

Also the work at \cite{wiedemann2020uhh} noted that fine-tuning BERT yield state of the art results for text classification tasks. Fine tuning BERT is typically done on task specific datasets in a supervised manner, it can also be done in an unsupervised manner by the masked language modeling MLM. In-domain unsupervised MLM pretraining allows for domain adaptation of the model. The authors concluded that since there is no way that the 2020 weakly labeled data could carry useful information to a machine learning system, they decided to use the manually labeled 2019 dataset only. However 2020 dataset was an ample source to build models using unsupervised learning particularly for domain adaptation using MLM training. The problem is that training a transformer based language model in an unsupervised manner is incredibly resource consuming making it impractical to learn from large datasets without access to larger GPU cluster or TPU hardware.

The authors at \cite{husseinnlpipassau} utilize the C-BiGRU model which is a CNN along with a bidirectional RNN. For work representation, a FastText is used to get the embeddings. They have participated in sub-task A of offenseval 2020 for the English. Turkish and Danish languages. For the base model the BERTbase model is used to get the embedding representation of each tweet so that it can be used for the classification task. The authors noted that fastText embedding performed significantly better than word2vec. That was because fastText used either Skipgram or CBOW mechanisms which generate vector representation for tokens that have not appeared in its training corpus. This behaviour is happening since it adds the character n-grams of all the n-grams representations i.e each word is treated as a collection of its constituent n-grams.  The authors provided key points for future improvements of the model. They suggested linking between tweets and their comment tweet such that we capture the context and the domain of tweet in a more coherent way since comment tweet might lose context without the main tweet. Furthermore, the author proposed used better representation of unknown tokens by using a matrix of embedding for all possible n-grams that can be extracted from the tokens, which can easily be extracted  form a fastText model representation, and that ultimately enhance the quality of words that have not been seen in the training set. Finally, the authors planned to further explore figurative language which is relatively hard to detect, this can be done be exploring the rhetorical features in the text \cite{caselli-lrec-2020}.

\begin{figure}[h]
	\centering
	\includegraphics[width=15cm]{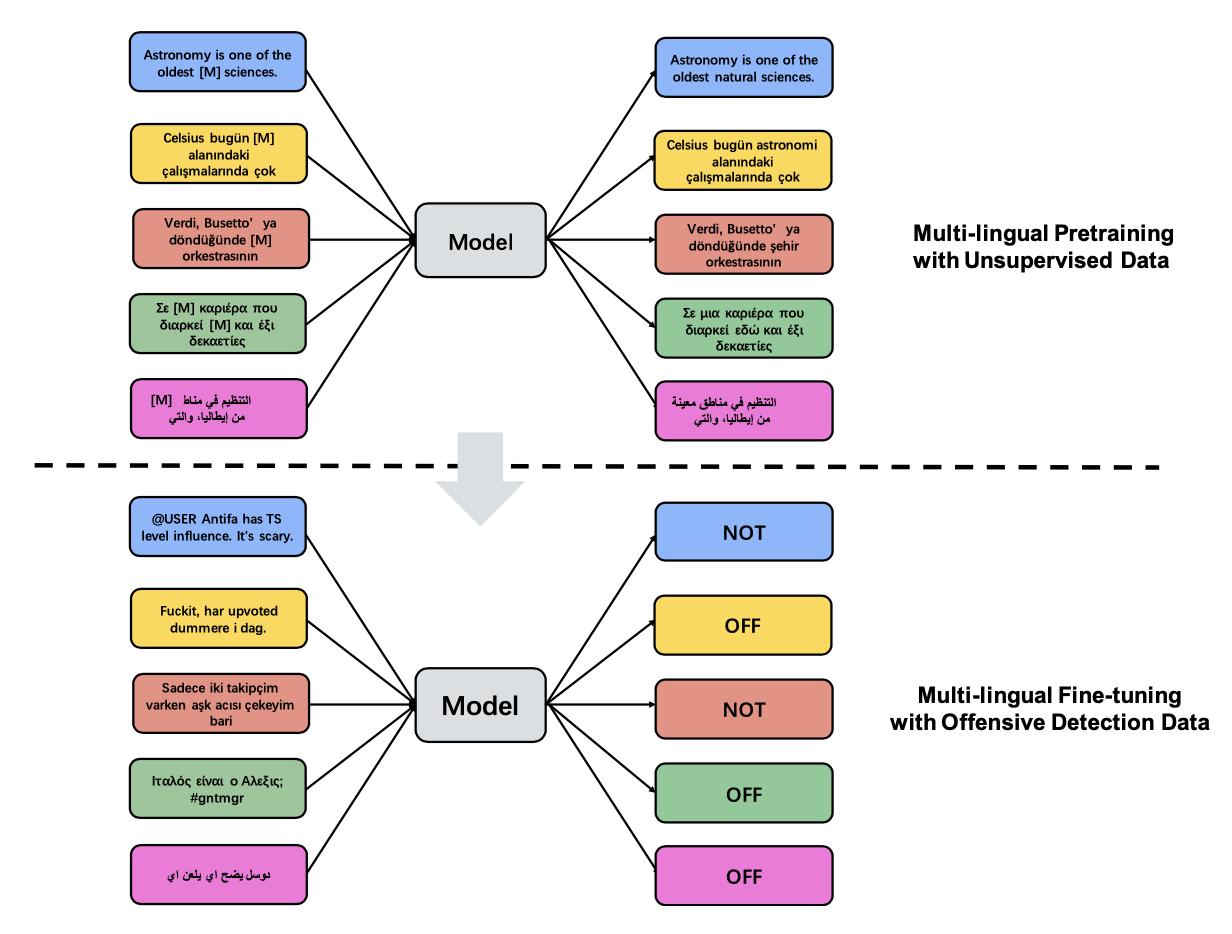}
	\caption{Pre-training and Fine-tuning Multilingual Model Paradigm}
	\label{fig:mulitlingualparadigm}
\end{figure}

Authors at \cite{wang2020galileo} built a unified approach to detect offensive utterances in all languages. The developed model involving two steps, first, pre-training using large scale multilingual unsupervised texts that yields a unified pre-trained model capable of learning all languages representations. The second step is to fine tune the pretrained model with labelded data. Since the pre-training step is very time consuming, the authors used an existing open source pretrained model XLM-R for the first step which was trained on over 100 language. Finally adding a fully connected layer for classification based on the [CLS] token of the top layer of XLM-R. Figure \ref{fig:mulitlingualparadigm} shows the two steps in the proposed model paradigm.

 Finally, in our participation at OffensEval 2020 \cite{hamdy2020nlpup} we have aimed to tackle the problem of detecting offensive language while considering both accuracy and model performance which is a major factor to deploying the model in a production environment. Throughout this work we have utilized an efficient implementation of FastText called BlazingText \cite{gupta2017blazingtext} and trained a model with a large dataset of more than 9 million tweets. The results obtained from blazing text were very competent to the much more advanced and large BERT based models while achieving a superior speed in training and making inferences.
 




\chapter{Neural Networks For Offensive Language Detection}\label{chap:Neural Networks For Offensive Language Detection}

In this chapter we are going to explain the different models used for hatespeech detection to discover why different models performs than others. Also we are going to perform an ablation to select the best preprocessing techniques and architecture for our task. This work depends on applying \textbf{BERT} and its efficient \textbf{RoBERTa} variation on the problem of Offensive language detection. Also, we will be using \textbf{BlazingText} which is a simple but very fast and efficient model based on fastText. We will also discuss the usage of a multilingual BERT model for Offensive language detection called \textbf{XLM-RoBERTa}. In the next chapter, a detailed comparison of the results will be presented.


\section{Transfer Learning and the offensive language detection}
Pretraining and fine tuning have become a new paradigm in NLP. In fact they represent together the notion of \emph{Transfer learning}, where general knowledge is initially acquired from a very large corpus and then transferred to down-stream tasks. 

Transfer learning has three main advantages; 1) Faster and cheaper model development: starting with a pre-trained model such as BERT already encodes a vast amount about the language trained on it, consequently, significantly less time training or fine-tuning for specific task. For instance BERT authors recommended only 2 to 4 epochs of fine tuning for specific task, where the most expensive of pre-training part is reuse. 2) Less data required: when training neural networks from scratch, a fairly large amount of data is required for the model to learn useful things and outperform traditional statistical methods. Transfer learning allows preserving the pre-trained model weights for the specific tasks which leads to the third advantage. 3) Obtaining better results: by simply fine tuning the preserved weights of the pre-trained model rather than training a custom model from scratch we can harness the knowledge acquired in the pretraining process and obtain state of the art results for our specific tasks.

A few prominent examples of Models that follow this paradigm include, ELMo and GPT for context aware word embeddings, BERT which propose a bidirectional language model, RoBERTa which customized BERT to removed the next prediction task and pre-train longer to get a better pre-trained model. In the next sections we have going to discuss the BERT and RoBERTa model in details and how they are applied to the offensive language detection problem.

With the lack of high quality dataset for offensive language detection, transfer learning based models such as BERT promises a huge advantages to efficiently detect and categorize hatespeech without the need of a large datasets. During the next section, we are going to discuss the basis of the BERT model, how the inputs are processed for fine-tuning and discussing the resulting output layers 

\section{Basis of the BERT model}
\begin{figure}[h]
	\centering
	\includegraphics[width=7cm]{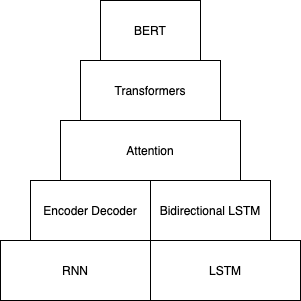}
	\caption{Concepts that sets the inspired BERT and set its basis}
	\label{fig:concepts}
\end{figure}

BERT architecture has been build and inspired by other prominent architectures preceding it. Figure \ref{fig:concepts} shows the concept hierarchy that inspired BERT, we will brief discuss each concept in this section.

\subsection{Recurrent Neural Networks (RNNs)}

Starting with the core architecture of Recurrent Neural Networks (RNNs) which was the base for processing a sequence of textual information for varies NLP tasks. Figure  \ref{fig:basicrnn} shows an overview of a basic RNN:

\begin{figure}[h]
	\centering
	\includegraphics[width=10cm]{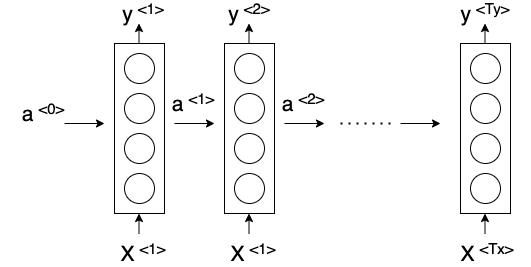}
	\caption{Basic RNN architecture based on many to many architecture for Named Entity Recognition}
	\label{fig:basicrnn}
\end{figure}

\begin{figure}[h]
	\centering
	\includegraphics[width=15cm]{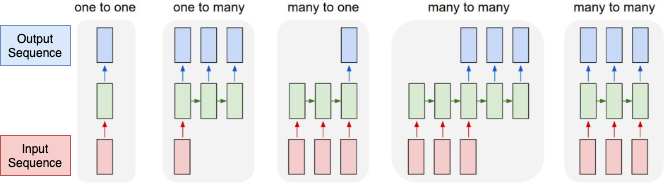}
	\caption{RNN arhitecture varients}
	\label{fig:rnnarchs}
\end{figure}

RNNs incorporate varies varient architectures that are suitable for a repective task including; many-to-many RNN for Named Entity Recognition and machine translation tasks; one-to-many RNN for music generation; many-to-one RNN for sentiment analysis tasks. Figure \ref{fig:rnnarchs} shows a nice visual illustration of these varies architecture variants published at Andrej Karpathy Blog \footnote{The Unreasonable Effectiveness of Recurrent Neural Networks. http://karpathy.github.io/2015/05/21/rnn-effectiveness/}. 

\subsection{Encoder and Decoder}
\begin{figure}[h]
	\centering
	\includegraphics[width=12cm]{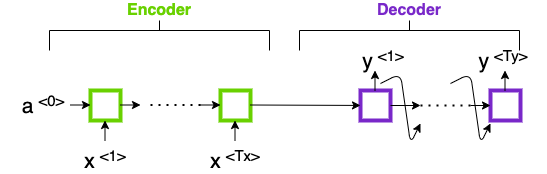}
	\caption{Encoder Decoder Architecture}
	\label{fig:encdec}
\end{figure}

BERT got inspired by a variant of the many-to-many RNN architecture namely the encoder and decoder architecture, where the encoder encode an input sequence trying to collect all the information into a final hidden state before passing it to the decoder. The final hidden state of the encoder is the initial state of the decoder to begin predicting the sequence. Figure \ref{fig:encdec} shows the encoder-decoder architecture.

\subsection{Long short-term memory (LSTM)}
The authors at \cite{sak2014long} introduced the LSTM RNNs to resolve the long term dependency problem of the vanilla RNNs when working with long text sequences specifically because of the vanishing gradients problem. A common LSTM unit is composed of a memory cell, an input gate, an output gate and a forget gate. The cell remembers values over arbitrary time intervals and the three gates regulate the flow of information into and out of the cell. Figure \ref{fig:lstmcell} shows the computation sequence inside LSTM cell.

\begin{figure}[H]
	\centering
	\includegraphics[width=9cm]{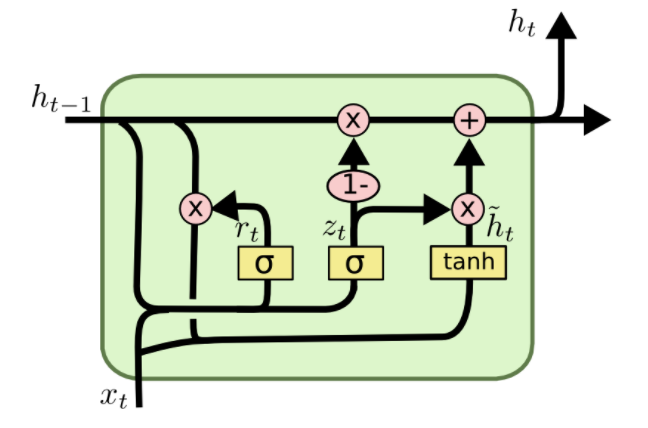}
	\caption{LSTML Cell Illustration}
	\label{fig:lstmcell}
\end{figure}

\subsection{Biredirectional RNN/LSTM}
Biderctional RNN introduced by \cite{schuster1997bidirectional} was an idea to build even more powerful sequence model. Birdirectional RNN enables the model to take information from both earlier and later in the sequence thus deeper understanding of the context. 

\begin{figure}[H]
	\centering
	\includegraphics[width=12cm]{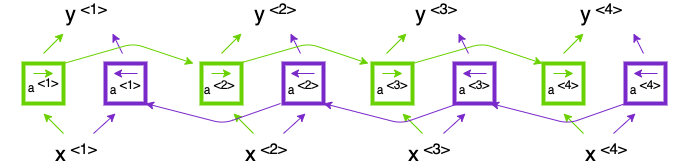}
	\caption{Birdirectional RNN architecture}
	\label{fig:lstmcell}
\end{figure}

\subsection{Attention and transformers}
\begin{figure}[h]
	\centering
	\includegraphics[width=14cm]{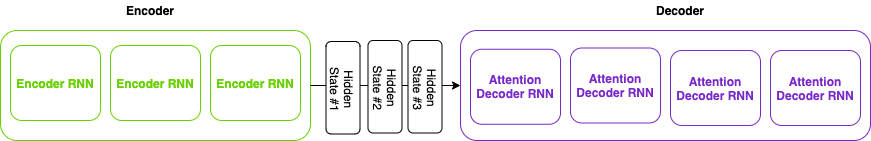}
	\caption{RNN architecture with Attention}
	\label{fig:attentionarch}
\end{figure}

The attention model is a modification of the encoder and decoder RNN architecture. The context vector becomes a bottleneck when working with long sequences. The authors at \cite{luong2015effective} introduced the attention model to tackle long sequences by only focusing on important parts of the input sequence. To accomplish that the encoder sends all the hidden state vectors to the decoder instead of passing the last hidden state as depicted in figure \ref{fig:attentionarch}. The decoder can now give attention to the relevant states from the encoder by giving it a score when processing the output in the corresponding decoding step. Then multiplying each hidden state with the softmaxed score to amplify or downing each state depending on relevancy score and finally summing up all the weighted vectors.

\begin{figure}[h]
	\centering
	\includegraphics[width=8cm]{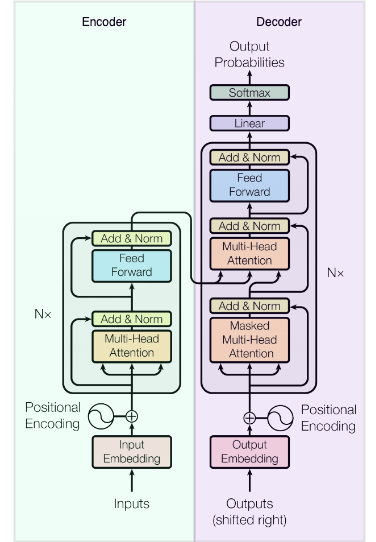}
	\caption{The Transformer model architecture}
	\label{fig:transformer}
\end{figure}

The attentions solved the long sequence bottleneck of the RNN model, however, RNN models also affected by its sequential nature and inability for implementing parallelization. To solve this issue, the Transformer is proposed by the authors at \cite{vaswani2017attention} to address this shortcoming of RNNs and taking advantage of the attention mechanism. Figure \ref{fig:transformer} shows the proposed transformer model architectures as depicted in \cite{vaswani2017attention}.

\section{The BERT model}
BERT is a pretrained model based on the Transformer model architecture. It shows states of the art results on many machine learning tasks. In the following subsections, we will be discuss the BERT model in details. We will start with how inputs are prepared and processed by the model, then we will cover the two tasks used for pretraining BERT, afterwards, a discussion on fine-tuning BERT for specific domains and the output of the BERT model. Finally we will present different variations of BERT that will be used in our experiments.

\subsection{Preparing inputs for BERT}
In this section we are going to discuss how to prepare the the tweets or text data for the BERT model digest. It is possible to fine-tune BERT and making predictions on any arbitrary text with unknown vocabularies in the dataset are included in the 30,000 words of the BERT vocabulary via tokenization. 

\subsubsection{BERT Tokenization}
Pre-trained BERT model comes with a build in tokenizer designed to take raw input text (as strings of characters) and transform it into tokens suitable for BERT input layer. BERT provides its own tokenizer since it has its own fixed set of vocabulary of tokens associated with a pre-trained embedding vector for each build-in token. One disadvantage of pretrained BERT is that it is not possible to used custom word embeddings. While all the knowledge and language understanding are available only on the embeddings that BERT pre-trained with, it is still shows state-of-the art with any arbitrary unknown words. The main reason for this outstanding performance is the intelligent method the BERT used to handle those out of vocabulary words. Also the existence of domain specific pretrained model that uses specialized corpora in the domain such as SciBERT trained on scientific literature and BioBERT which is trained on huge corpus of medical text.

BERT tokenizer relies on \textbf{WordPiece model} to tokenize the textual inputs. Amount the 30,000 vocabulary in BERT, 80 percent of them are whole words while the remaining are subwords. The way BERT handle out-of-vocab words is to breakdown them into subwords, in case of missing subwords, it further breakdown unknown subwords into individual character until BERT can come up with a reasonable representation of the meaning of the whole unknown word.  Along side individual characters, BERT also has punctuation characters which in some cases contribute to the general understanding of sentences while only the whitespace are discarded by BERT tokenizer. The main reason BERT on having small vocabulary of common forms of words is that including rare words will results on poor representations since they are limited withing any given dataset while using subwords for representation would held better relationships between the different forms of a word. Following the tokenization there is a numericalization step that maps each token to a unique integer in the corpus’ vocabulary for looking up the embedding of each vector.

\subsubsection{Special tokens in BERT}
BERT also include some special tokens for its inputs which they are:
\begin{enumerate}
\item The classification \textbf{[CLS]} token for sentence level classification tasks. The token is always added to the start of every sentence.
\item The separator \textbf{[SEP]} token for segment embeddings of the Next Sentence Prediction task.
\item The unknow \textbf{[UNK]} token for replacing vocabulary that was not found in BERT words or subwords.
\item The padding \textbf{[PAD]} token for padding short sentence to make sure that all input sequences in the batch are of the same length when the embedding are stacked together.
\end{enumerate}

The following example sentence shows how BERT tokenizer works:

\begin{figure}[H]
	\centering
	\includegraphics[width=15 cm]{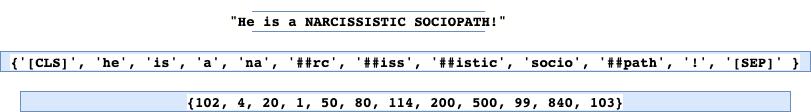}
	\label{fig:tokenizedsenexample}
\end{figure}

\subsubsection{BERT processing parallelization and handling variable sentence length}
For each input token embedding, BERT model consider all other token embedding for processing to get the context of the word and produce a better embedding for the respective token. That way can handle the processing for all tokens in parallel since that processing each word is independent from other words. That also enables BERT to handle input sentences of different lengths while this would be handled in recurrent neural network by sequentially processing each work one after the other. The process of taking a set of token embedding and producing better ones are repeated 12 times across the 12 layers of BERT. Figure \ref{fig:bertarch} shows the 12 layers of BERT architecture and how the input for each layer:

\begin{figure}[H]
	\centering
	\includegraphics[width=15cm]{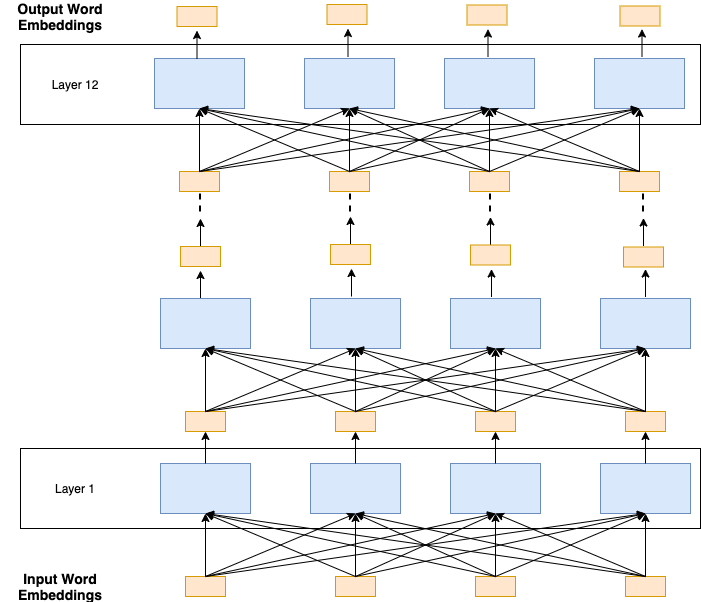}
	\caption{Input processing across the 12 layers of BERT}
	\label{fig:bertarch}
\end{figure}

\subsection{Positional encoding and segment embeddings}
In RNN, the order in which the word appears in the sentence is crucial since the are fed in to the network one at a time in a sequential order, however BERT does not have dependency of the order of the input words. To make sense of the word order, BERT incorporate an layer of embeddings called \textbf{positional embeddings PE}. For each token in the sentence, the model is fed the vector embedding and added to it the \textbf{PE} vector corresponding the token position thus understanding the relative position of the words but not only the absolute position. Also a segment embedding for each layer has to be added to handle two-sentence tasks. Figure \ref{fig:bertinputs} shows the inputs to the BERT first layer:

\begin{figure}[H]
	\centering
	\includegraphics[width=15cm]{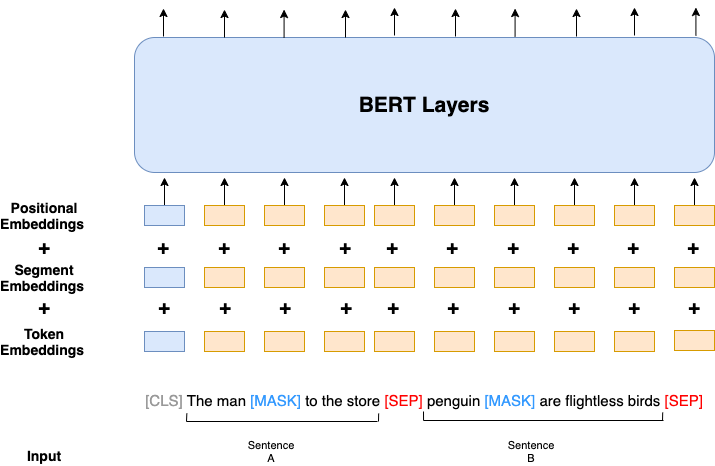}
	\caption{Input embeddings to BERT}
	\label{fig:bertinputs}
\end{figure}

For each extracted token BERT tokenizer look up the embeddings which are going to be the actual inputs to the BERT model. Each embedding is a vector of 768 features. Looking at BERT as a black box, BERT's mission is to take each embedding and outputs and enhanced version of it with the exact same dimension.

\subsubsection{BERT inner working}
BERT architecture mainly consisting of the encoder part of the Transformer model. The encoder architecture of the Transformers uses a mechanism called Self Attention to help the model learn the context of the sentence. Self attention is the mechanism to allow Transformers to look at other words embeddings similar to how RNN maintain the hidden states of previous words in the chain when processing the current word. 

\begin{figure}[H]
	\centering
	\includegraphics[width=15cm]{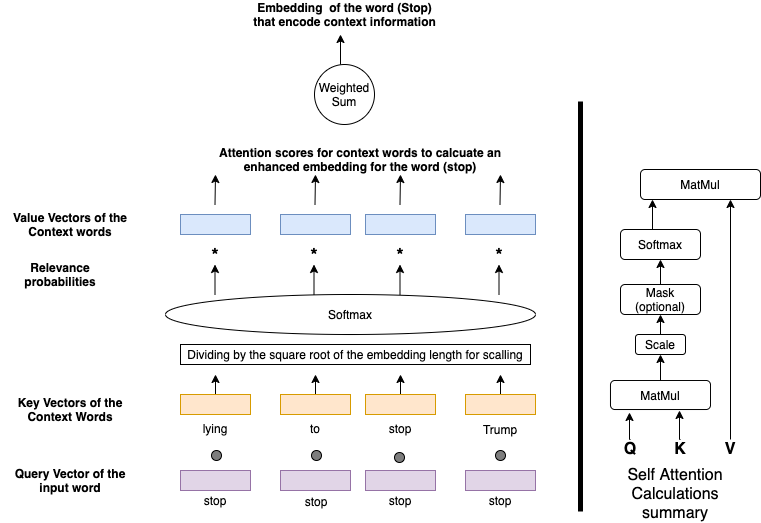}
	\caption{Self attention mechanism overview}
	\label{fig:selfattention}
\end{figure}

To calculate the attention score, first the context words embeddings are projected into three vector spaces by multiplying them into learned weights, the \textbf{Query space} weights to get the query vectors, the \textbf{Key spaces} for the key vectors and finally the \textbf{Value space} for value vectors. First step, the dot product of the query vector of the target word with all the context key words, producing a score. Then applying softmax activation function to the produced scores from the first step. And multiplying the value vectors of the context words by the softmax score. The final results is the attention score for each word that is a representation of the influence of the context word on a the target word. An example of self attention calculation is depicted in figure \ref{fig:selfattention}

The above self attention diagram, showing the calculation to extract the embedding of a single word in the input sentence, notably this calculations can be parallalized for all the words in a sentence to calculate the embedding that encode information about the order and importance of other context words to the target word.

BERT incorporate the idea of Self Attention with multiple instances, each instance of Self Attention is called Attention Head. BERT architecture consists of 12 attention head, each of which has its own unique instance of the Q, K, and V projection matrices. Each of the attention head calculates an embedding for the target word and finally all the embedding from the different heads and concatentated and normalized as shown in figure \ref{fig:multihead}

\begin{figure}[H]
	\centering
	\includegraphics[width=15cm]{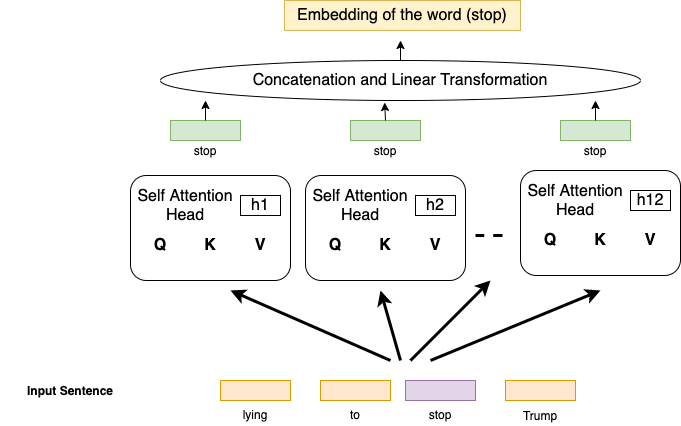}
	\caption{Multi Head Attention overview}
	\label{fig:multihead}
\end{figure}

For every layer of the BERT model, the inputs are passed throght a multihead attention layer and then through a feed forward neural network before it gets feed to the next layer. The illustration \ref{fig:basicrnn} shows the overview of a single BERT layer:

\begin{figure}[H]
	\centering
	\includegraphics[width=10cm]{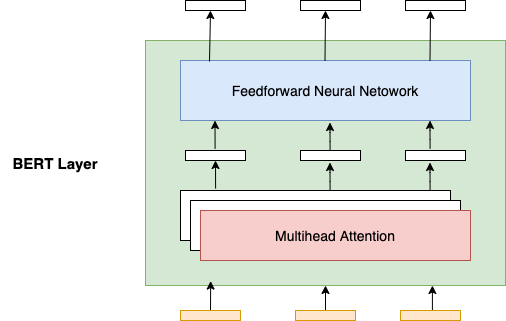}
	\caption{Overview of the component of a single BERT layer}
	\label{fig:multihead}
\end{figure}

\subsection{Variations of BERT model}
After the BERT model has been released to the public, many researchers introduced varies enhancements to the BERT model to push forward the potential of the BERT architecture. These enhancements where either concerned about updating different aspects of the BERT architecture, experimenting with different hyperparameters, adjusting the training objectives, the type of training dataset  and the length of the pretraining process. In this work, we have experiments with top performing models variants and applying them to the OLD dataset to classify offensive language while applying the necessary proprocessing and model adjustments to optimize the results.

\subsubsection{ALBERT - A lite BERT for self-supervised learning}
The authors at \cite{lan2019albert} proposed ALBERT as a salable architecture for BERT. While BERT architecture can easily be adjusted to increase the number of layer (making it more deep) and also increasing the model width, it wasn't salable. According to the authors, enlarging the BERT architecture beyound 12 layers and 1024 hidden units diminishes the accuracy on the task. 

In ALBERT is based on the BERT with changes in its architecture and training process. The changes are mainly conducted with two main goals. The first is to improve the speed of the pre-training at all size variants, and greatly improve the results of the large models which was either time consuming and inefficient in the original BERT architecture. These changes are as following:
\begin{itemize}
    \item \textbf{Parameter sharing}: instead of having unique parameters for each layers of the BERT model, ALBERT uses identical parameters for all its twelve layers. This can be thought of as using a single encoder layer that applies twelve time to the input embedding. There are few effects of parameter sharing that can be summarized as following:
    \begin{itemize}
        \item It has\textbf{ increased the training speed}, by sharing the weights it reduced the number of unique parameters in the model and therefore faster backpropagation. However, the inferencing process didn't see much improvements since it requires the same number of computation for the forward pass.
        \item The effect on accuracy was insignificant with a minor decrease for the same number of layers. Hoewever, with a scalable architecture, adding more layers improves the results over BERT.
        \item Enables a more scalable architecture with the reduced footprint as a result of parameter sharing. We can now add more layers without affecting the number of parameters for faster training and effective model outcome. The table \ref{} from the ALBERT paper shows the configurations of ALBERT compared to BERT.
    \end{itemize} 
    

    \item \textbf{Factorizing the Embeddings:} Using a smaller word embedding with only 128 features compared to the original BERT of 786 features that reduces the number of parameters from 23 million to only 3.8 million parameters. ALBERT uses a projection matrix to scale the embedding back to 768 features. ALBERT author noted that with the achieved 80\% reduction in the parameter of the embedding layer, there is only a minor drop in performance on the benchmark dataset with all other parts identical to BERT. 
    \item \textbf{Large Batch Optimization (LAMP optimizer) for pre-training:} instead of Adam optimizer, ALBERT uses LAMP optimized to train with large patches for better utilization of the GPU while addressing the unstable convergence while using large learning rates.  
    \item \textbf{Sentence Order Prediction SOP:} While the original BERT model involves the Next Sentence Prediction task, ALBERT authors recognized that using Sentence Order Prediction task is a better alternative as a training objective compared to NSP which overlaps with what is already learned in the MLM tasks . SOP focuses on modeling inter-sentence coherence.
    \item \textbf{N-gram Masking} Similar to BERT, ALBERT is also trained on a masked language model tasks. However, the only distinction from BERT is that it chooses a contiguous sentences of tokens refered to as n-grams instead of randomly picking 15\% of the sentence to be masked.
\end{itemize}

\subsubsection{RoBERTa - A Robustly Optimized BERT Pretraining Approach}
RoBERTa \cite{liu2019roberta} is a variation of the BERT model where the authors evaluated the BERT model for the most effective hyperparameters and most suitable training set size. This lead to four main modifications to improve BERT performance:
\begin{itemize}
    \item Pre-training the model for longer period with bigger batches over the data.
    \item Removing the next sentence prediction NSP task.
    \item Training on longer text sequences.
    \item Dynamically changing the masking pattern applied to the training data.
\end{itemize}
		
The core BERT architecture is fixed in RoBERTa while offering an enhanced recipe for training BERT including updated model hyperparameter value that proved better performance for the resulting pretrained model. RoBERTa surpasses the results from BERT on benchmark datasets including GLUE, RACE and SQuAD.

\subsubsection{Multilingual offensive Language detection with XLM-RoBERT}
The idea behind Multilingual BERT is to pretrain it using as many language as long as we can get a substantial amount of text for it. Then feeding all the data from all language as they were all from the same language, thus treating all different language as one language with a very large character set and vocabulary.XLM-

\subsection{Fine Tuning methods of BERT based models}
Pretraining language model proved efficiency to learn universal language representation that can be integrated for various NLP tasks while avoiding training model from scratch. Text classification is one of the most common problems in NLP, the main goal is to assign predefined categories to a given document. There are substantial work that have proved the benefits of pre-training model including word embedding such as word2vex and GloVe; Contextual word embedding such as CoVe and ELMo. Another knind of pre-training model is sentence level such as ULMFiT that achieved state of the art results on text classification tasks. And more recently OpenAI GPT and BERT. Three solutions to fine-tune pretrained BERT model to improve performance on classification tasks. 

\begin{enumerate}
\item \textbf{Further pre-train BERT within-task training data or in-domain data}: While the BERT team offered a pre-trained model based on large amounts of general text data, BERT implementation offers the capability to pre-train the model from scratch. The decision of pretraining the model from scratch might be taken if task under development doesn't benefit from the BERT understanding of general text on which it was trained on. However in the majority of tasks, it appears that training on general text enabled BERT to achieve state of the art performance on tasks with different context. Other domain specific pretrained models such as \textbf{SciBERT} based on biomedical and computer science literature corpus, \textbf{FinBERT} based on financial services corpus), \textbf{BioBERT} based on biomedical literature corpus, \textbf{ClinicalBERT} based on clinical notes corpus, \textbf{mBERT} based on corpora from multiple languages and patentBERT based on patent corpus. The authors at \cite{caselli2020hatebert} also proposed the hateBERT variant where they pretrained the BERT model on hate speech text which was shown to perform better on offensive language detection tasks compared to models pretrained on general or scientific text.

\item \textbf{Fine tune Bert for the target task}: The most common approach of using BERT is fine-tuning a model that has been trained with massive amounts of data. This approach is particularly popular since the process is timely feasible for most practitioners where the fine tuning can take minutes to very few hours for a fairly sized data set for the task. The fine-tuning process involves updating the pretrained weights of BERT layers. It is also possible to target specific layers for fine-tuning while freezing the weights of other layers \cite{sun2019fine}. Our approach involving fine-tuning all the layer and a shallow neural network for classification to get the best results possible for the training. 
`                                                   `   1
\item \textbf{Fine tune Bert with multiple task learning if several related tasks are available}: training BERT based on multiple tasks simultaneously without training a general language model. This approach showed efficiency to leverage the shared knowledge among the different tasks.

\item Using BERT to extract fixed feature vectors: Instead of end-to-end pretraining the BERT model, it can be extremely useful to utilize BERT as ELMo to extract high quality contextual representations/embeddings based on the task dataset that can be used in a machine learning pipeline for building a model. 
\end{enumerate}


\section{BlazingText for fast predictions}
BlazingText was introduced by researchers at Amazon \cite{gupta2017blazingtext}. They propose an efficient implementation of Word2Vec \cite{mikolov2013distributed} and FastText classifier \cite{joulin2016bag} that can take advantage of multiple CPUs and GPUs in training the model and real-time inference. Amazon offers an implemetation of the BlazingText algorthms provided via their AWS SegaMaker managed service for data science. SegaMaker allows for complete datascience workflow on the cloud, by providing Python Jupyter notebook for developing the model while efficiently managing preprocessing, training and inferencing capabilities by off-loading these workloads to user defined GPU instances and charging the user for the exact running time per second precision of the compute instances. 

We have also turned on automatic hyperparameter tuning for the learning rate, vector dimension, minimum word count and word n-grams parameter of the BlazingText for classification model.


\chapter{Experiments setup and Results}\label{chap:results}

This chapter discusses the experiments details that were conducted to fulfil the research goals of this thesis. The first sections presents the overall and general setup for the experiments including the general processes that were followed on all the experiments and the performance metrics. The next section, we will provide the details for each experiments to ensure ease of reproducibility and finally the detailed outcome of each experiments and comparison of the results.

\section{Datasets}

Our experiments are conducted based on the offensive language detection tasks of the OffenseEval challenges in both the 2019 and 2020 rounds of the challenge.

The OffensEval conference provided researchers in the field of Offensive language detection with a valuable dataset with high potential for varies experimental setup. In 2019 the OLD dataset has been introduced by the authors at \cite{zampieri2019predicting} which was was used as a bases for the SemEval Task 6 for offensive language detection.

The 2019 OLD dataset contain tweet text collected from Twitter using the official API composed of around 13240 tweets. The authors used a hierarchical annotation schema to spit the dataset into three levels to distinguish: 
\begin{itemize}
\item (Task A) whether language is offensive or not. There are 8840 (66.8\%) examples of non offensive tweets and 4400 (33.2\%) instance of offensive tweets.
\item (Task B) the type of offensive language. The number of offensive tweets with targeted intent is 3876 tweet while there are 524 tweet of untargeted tweets.
\item (Task C) the offense target identification. There are 1074 instances of the offensive tweets target a group, 2407 that targets an individual, and 395 that target other entities.
\end{itemize}

\begin{figure}[H]
	\centering
	\includegraphics[width=15cm]{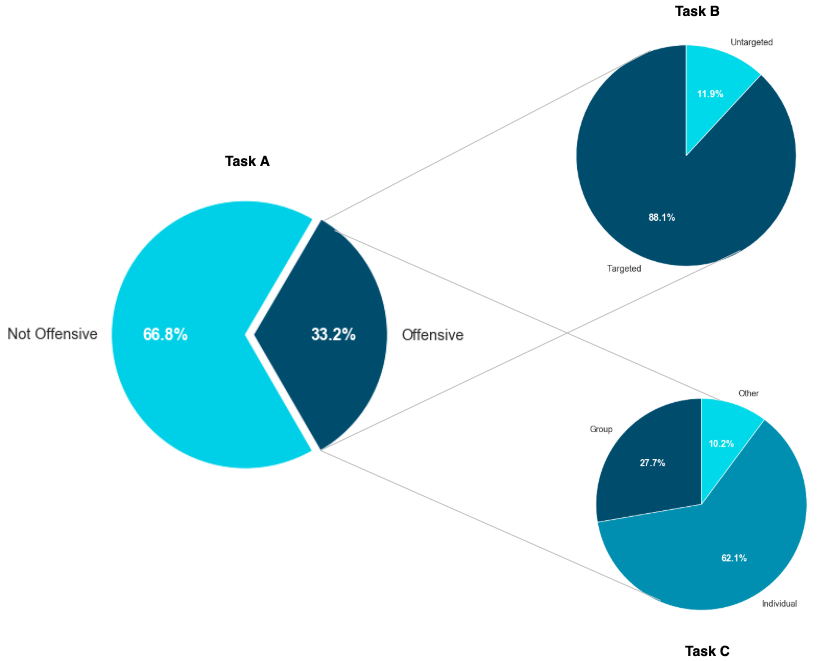}
	\caption{Class Distribution of the OLD 2019 dataset}
	\label{fig:old2019}
\end{figure}

The dataset has been manually annotated by expert annotators and therefore the 2019 dataset represents one of the highest quality dataset for offensive language detection. Figure \ref{fig:old2019} shows the distribution of the three level categories and the number of example in each category.

\begin{figure}[H] 
	\centering
	\includegraphics[width=10cm]{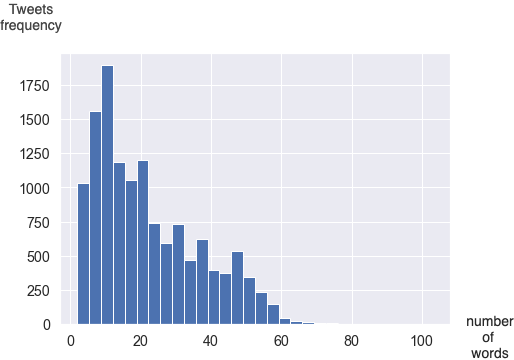}
	\caption{Sentence length distribution of the OLD 2019 dataset}
	\label{fig:dist2019}
\end{figure}

The majority of tweets in the 2019 OLD dataset are between 5 to a maximum of 60 words. Determining the sentence length is an important decision when preparing the input data for BERT based models. Figure \ref{fig:dist2019} shows the distribution of the tweets length across the whole dataset.

\begin{figure}[H]
	\centering
	\includegraphics[width=15cm]{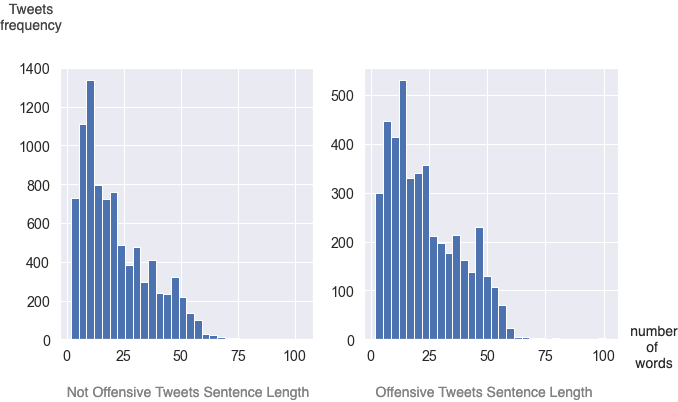}
	\caption{Sentence length distribution per class of the OLD 2019 dataset}
	\label{fig:dist2019class}
\end{figure}

To show if there is a difference between the sentence lengths of offensive versus non offensive tweets, we have also depicted the tweets length distribution for the respective class in Figure \ref{fig:dist2019class} and they show similar word length distribution with a maximum sentence length of around 60 words.

After the wide interest of the NLP community in the offensive language detection task in SemEval2019, they have offered another challenge for detecting offensive language based on a semi-supervise dataset SOLID 2020 \cite{zampieri2020semeval}. The dataset was available on multiple languages including Arabic, Danish, English, Greek, and Turkish. The focus in this these will be mainly on the English language dataset to conduct the experiments. 

SOLID follows the same hierarchical modeling of offensive language as in OLD 2019 discussed above. However, the dataset has been different annotated. Compared to 2019 dataset, the 2020 English dataset has been annotated in a semi supervised manner using democratic co-trainining and OLID as a seed dataset. This allowed for the collection of a massive number of training annotated example which accounts to over 9 million examples on the cost of accuracy. However the datasets for the other languages (Arabic, Dutch, Greek, and Turkish) were manually annotated by volunteer annotators and native speakers. Similarly to the OLID 2019, the SOLID 2020 resembles the same class distribution where there are some classes where underrepresented.



\section{Preprocessing}
For all the experiments in this chapter, there are few simple but rather effective preprocessing steps that has been applied to clean the dataset and enrich the sentences context. These preprocessing steps are as following:

\begin{itemize}
\item Removing user mentions to reduce redundancy and unnecessary data volume that would negatively affect the training runtime and performance.
\item Replace emoji Unicode with their representative text to enhance the sentence context information since most social media users express their feelings using emojis.
\item The words in hashtags are retained with only the removal of the \# sign.
\item Removing numbers of special characters.
\item Applying sentence segmentation to separate concatenated words in a sentence.
\end{itemize}

\subsection{Formatting inputs for PlazingText:}
The BlazingText algorithm is trained based on a training and validation dataset, each of which is expected to be in a dedicated file. Each line in the dataset file contains a single example. The label of each example must precede the example sentence and prefixed by the string "\_\_label\_\_". Here are few example sentences from the processed training file. 

\begin{verbatim}
__label__0 obama wanted liberals and illegals to move into red states
__label__1 liberals are all kookoo
__label__0 i'm so fucking ready
__label__1 canada doesn need another cuck we already have enough
\end{verbatim}
				
For out testing purposes we have used the batch transfer mode of the trained blazingText model to get inferences for all the example in the test dataset. The test dataset is expected to be preprocessed into JSON lines format. Here are some example of the test dataset:

\begin{verbatim}
{"source": " user  me too are all racist "}
{"source": " liberal logic if  liberals get their way url"}
{"source": " sierra burgess is a loser she is me when
            my phone dings  face with tears of joy "}
\end{verbatim}

\section{Experiment Setup}
This section describes the implementation details of the training process for our experiments such as the dataset splitting, hyperparameters selection, the hardware used and the evaluation metrics to compare the performance of each model. 

\begin{figure}[H]
	\centering
	\includegraphics[width=15cm]{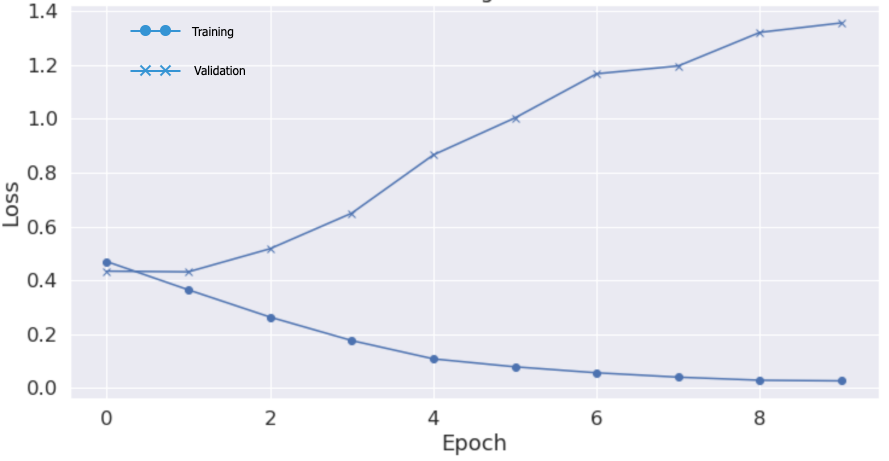}
	\caption{Loss updates on the training and validation dataset across 10 epochs}
	\label{fig:trainvalidloss}
\end{figure}

Transformers based models, we have utilized some of the top performing transformer based models including BERT, RoBERTa and ALEBERT models based on the model rank on the benchmark leaderboards. We have used the pretrained model provided by the [huggingface] library. For all model we have first added a linear layer with RELU activation and a dropout layer of probability \textbf{0.1} to reduce overfitting on the training set and finally a LogSoftmax layer to get probability output for the two classes (offensive and not-offensive). A learning rate of \textbf{2e-5} that is recommended by the Google BERT team is used, it is small enough to ensure convergence on the training and validation  datasets. The model is trained for only \textbf{3} epochs to avoid the fine tuning process to overfit and the training set and negatively affects the results on the test dataset. Figure \ref{fig:trainvalidloss} shows the model tendency to overfit the training data and significantly worsen the results on the validation dataset. The training is done on a machine equipped with a Nvidia 2080Ti GPU and 16 GB of RAM. Training for 3 epochs takes 12 minutes to complete on the  the OLID 2019 development dataset which contains over 13000 records, the development dataset has been split with a ration of \textbf{80\%} as a training data and \textbf{20\%} as a validation dataset. We haven't performed cross validation due to the framework limitation. The final performance is evaluated based on the F1 scores which gives a better judge for an unbalanced dataset

For the BlazingText algorithm the experiment is completed on AWS Sagemaker using a \textbf{ml.c5.4xlarge} instance. The dataset for the OLID 2020 preprocessed as mentioned in the previous sections and store on AWS S3 bucket. With the massive size of the 2020 dataset which is over 9 million records we have used a splitting of \textbf{99\%} as training data and \textbf{1\%} as a validation dataset. Using hyperparameter tuning, the chosen learning rate was \textbf{0.05}, the model is set to be trained for \textbf{1}0 epochs with early stopping flag set to true which causes the training process to stop after the \textbf{7th} epoch since there is no performance gains after that point on the validation set. 

\section{Results}
This section we will preset the results of our experiments. The results from the two test datasets of offenseEval 2019 and 2020 will be presented separately. Our main focus is to target task A and B in the dataset. We will also specify which datza are used in the training process with respect to the model. For evaluation purposes, we mainly focus on the macro weight average \textit{F1-Score} for comparison reasons. The models are ordered in the tables based on the best F1 score achieved on the respective task.

There are five systems that we will show the results on each of the dataset and comparing it with the best system results in the OffensEval competition. For the base line model, we have trained a logistic regression model based on features extracted from TFIDF, Doc2Vec, 3-gram and embedding extracted from the BERT model.  

\subsection{Task A: Offensive or not}
In this task the goal is to classify tweets into two categories, either offensive or non offensive. Table \ref{tab:taskABaseResults} shows our baseline model trained based on varies features, we will be using the model with the highest F1 score as own standard baseline. In tables \ref{tab:taskAResults2020} and \ref{tab:taskAResults2019} and  we report the results on subtask a for all the methods we have used.

\begin{table}[h!]
\centering
\begin{tabular}{llllll}
\multirow{2}{*}{\textbf{Features}} & \multicolumn{2}{|c|}{\textbf{OLID2019}}                                & \multicolumn{2}{c}{\textbf{OLID2020}}                   \\
\cline{2-5}
                  & \multicolumn{1}{|c|}{Accuracy}      & \multicolumn{1}{c|}{F1}    & \multicolumn{1}{c|}{Accuracy}      & \multicolumn{1}{c}{F1}                \\
\hline
\textbf{BERT embeddings}         & \multicolumn{1}{|c|}{83.7\%} & \multicolumn{1}{|c|}{0.791} & \multicolumn{1}{|c|}{93\%} & \multicolumn{1}{c}{0.87}  \\
n-gram (range 1-3)         & \multicolumn{1}{|c|}{80.4\%} & \multicolumn{1}{|c|}{0.71} & \multicolumn{1}{|c|}{93.7\%} & \multicolumn{1}{c}{0.862}  \\

TFIDF         & \multicolumn{1}{|c|}{80.9\%} & \multicolumn{1}{|c|}{0.70} & \multicolumn{1}{|c|}{93.3\%} & \multicolumn{1}{c}{0.86}    \\
Doc2Vec         & \multicolumn{1}{|c|}{79.8\%} & \multicolumn{1}{|c|}{0.68} & \multicolumn{1}{|c|}{92.3\%} & \multicolumn{1}{c}{.85}      \\

\hline
\end{tabular}
\caption{Results of the logistic regression baseline model based on varies features for the OLD dataset Task A}.
\label{tab:taskABaseResults}
\end{table}

\begin{table}[h!]
\centering
\begin{tabular}{llllll|l}
\multirow{2}{*}{\textbf{Model}} & \multicolumn{2}{|c|}{\textbf{Accuracy}}                                & \multicolumn{2}{c|}{F1}                    &        \multirow{2}{*}{\textbf{Runtime}}     &     \multirow{2}{*}{\textbf{Training Data}}  \\
\cline{2-5}
                  & \multicolumn{1}{|c|}{CV}      & \multicolumn{1}{c|}{Gold}    & \multicolumn{1}{c|}{CV}      & \multicolumn{1}{c|}{Gold}      &            \\
\hline
\textbf{RoBERTa}         & \multicolumn{1}{|c|}{99.1\%} & \multicolumn{1}{|c|}{92.6\%} & \multicolumn{1}{|c|}{0.923} & \multicolumn{1}{c|}{0.913}        &     \multicolumn{1}{c|}{12 minutes}  &    OLID2019\\
BERT         & \multicolumn{1}{|c|}{98.2\%} & \multicolumn{1}{|c|}{92.3\%} & \multicolumn{1}{|c|}{0.919} & \multicolumn{1}{c|}{0.91}        &     \multicolumn{1}{c|}{12 minutes}      & OLID2019\\

\textbf{BlazingText}       & \multicolumn{1}{|c|}{96.89\%}     &  \multicolumn{1}{|c|}{92.1\%} & \multicolumn{1}{|c|}{0.92} & \multicolumn{1}{c|}{0.9088}   &  \multicolumn{1}{c|}{40 seconds} & OLID2019 + OLID2020\\

ALBERT         & \multicolumn{1}{|c|}{98.1\%} & \multicolumn{1}{|c|}{92.3\%} & \multicolumn{1}{|c|}{0.917} & \multicolumn{1}{c|}{0.902}        &     \multicolumn{1}{c|}{12 minutes}      & OLID2019\\
XLM-RoBERTa         & \multicolumn{1}{|c|}{97.16\%} & \multicolumn{1}{|c|}{92.3\%} & \multicolumn{1}{|c|}{0.905} & \multicolumn{1}{c|}{0.898}        &     \multicolumn{1}{c|}{12 minutes}     & OLID2019 \\

BaseMode         & \multicolumn{1}{|c|}{96.37\%} & \multicolumn{1}{|c|}{93\%} & \multicolumn{1}{|c|}{0.929} & \multicolumn{1}{c|}{0.87}        &     \multicolumn{1}{c|}{12 minutes}       & OLID2019\\

\hline
All NOT           &   \multicolumn{1}{|c|}{83.87\%}    & \multicolumn{1}{|c|}{72.3\%} & \multicolumn{1}{|c|}{0.459}  &  \multicolumn{1}{c|}{0.42}    &    \multicolumn{1}{c|}{NA}   &  \\
All OFF           &   \multicolumn{1}{|c|}{16.1\%}    & \multicolumn{1}{|c|}{27.7\%} & \multicolumn{1}{|c|}{0.138}   &  \multicolumn{1}{c|}{0.217}   &   \multicolumn{1}{c|}{NA}    &   \\
\hline
\end{tabular}
\caption{Experimental results of sub-task A (CV = cross-validation; gold = gold test set) on the 2020 test data set}.
\label{tab:taskAResults2020}
\end{table}

\begin{table}[h!]
\centering
\begin{tabular}{llllll|l}
\multirow{2}{*}{\textbf{Model}} & \multicolumn{2}{|c|}{\textbf{Accuracy}}                                & \multicolumn{2}{c|}{F1}                    &        \multirow{2}{*}{\textbf{Runtime}}     &     \multirow{2}{*}{\textbf{Training Data}}  \\
\cline{2-5}
                  & \multicolumn{1}{|c|}{CV}      & \multicolumn{1}{c|}{Gold}    & \multicolumn{1}{c|}{CV}      & \multicolumn{1}{c|}{Gold}      &            \\
\hline
\textbf{RoBERTa}         & \multicolumn{1}{|c|}{84.1\%} & \multicolumn{1}{|c|}{0.84\%} & \multicolumn{1}{|c|}{0.833} & \multicolumn{1}{c|}{0.81}        &     \multicolumn{1}{c|}{12 minutes}  &    OLID2019\\
ALBERT         & \multicolumn{1}{|c|}{84.1\%} & \multicolumn{1}{|c|}{84.1\%} & \multicolumn{1}{|c|}{0.837} & \multicolumn{1}{c|}{0.801}        &     \multicolumn{1}{c|}{12 minutes}      & OLID2019\\

BERT         & \multicolumn{1}{|c|}{83.5\%} & \multicolumn{1}{|c|}{83.13\%} & \multicolumn{1}{|c|}{0.795} & \multicolumn{1}{c|}{0.788}        &     \multicolumn{1}{c|}{12 minutes}      & OLID2019\\
XLM-RoBERTa         & \multicolumn{1}{|c|}{80.1\%} & \multicolumn{1}{|c|}{81.2\%} & \multicolumn{1}{|c|}{0.793} & \multicolumn{1}{c|}{0.781}        &     \multicolumn{1}{c|}{12 minutes}     & OLID2019 \\

\textbf{BlazingText}       & \multicolumn{1}{|c|}{84\%}     &  \multicolumn{1}{|c|}{85\%} & \multicolumn{1}{|c|}{0.812} & \multicolumn{1}{c|}{0.797}   &  \multicolumn{1}{c|}{40 seconds} & OLID2019 + OLID2020\\

BaseMode         & \multicolumn{1}{|c|}{85.8\%} & \multicolumn{1}{|c|}{81\%} & \multicolumn{1}{|c|}{0.837} & \multicolumn{1}{c|}{0.791}        &     \multicolumn{1}{c|}{12 minutes}       & OLID2019\\

\hline
All NOT           &   \multicolumn{1}{|c|}{83.87\%}    & \multicolumn{1}{|c|}{72.1\%} & \multicolumn{1}{|c|}{0.459}  &  \multicolumn{1}{c|}{0.42}    &    \multicolumn{1}{c|}{NA}   &  \\
All OFF           &   \multicolumn{1}{|c|}{16.1\%}    & \multicolumn{1}{|c|}{27.9\%} & \multicolumn{1}{|c|}{0.138\%}   &  \multicolumn{1}{c|}{0.217}   &   \multicolumn{1}{c|}{NA}    &   \\
\hline
\end{tabular}
\caption{Experimental results of sub-task A (CV = cross-validation; gold = gold test set) on the 2019 test data set}.
\label{tab:taskAResults2019}
\end{table}

\subsection{Task B: Offense types}

\begin{table}[H]
\centering
\begin{tabular}{llllll|l}
\multirow{2}{*}{\textbf{Model}} & \multicolumn{2}{|c|}{\textbf{Accuracy}}                                & \multicolumn{2}{c|}{\textbf{F1}}                    &        \multirow{2}{*}{\textbf{Runtime}}     &     \multirow{2}{*}{\textbf{Training Data}}  \\
\cline{2-5}
                  & \multicolumn{1}{|c|}{CV}      & \multicolumn{1}{c|}{Gold}    & \multicolumn{1}{c|}{CV}      & \multicolumn{1}{c|}{Gold}      &            \\
\hline
\textbf{RoBERTa}         & \multicolumn{1}{|c|}{78\%} & \multicolumn{1}{|c|}{74.5\%} & \multicolumn{1}{|c|}{0.748} & \multicolumn{1}{c|}{0.732}        &     \multicolumn{1}{c|}{12 minutes}  &    OLID2019\\
XLM-RoBERTa         & \multicolumn{1}{|c|}{72.2} & \multicolumn{1}{|c|}{70.6\%} & \multicolumn{1}{|c|}{0.71} & \multicolumn{1}{c|}{0.69.6}        &     \multicolumn{1}{c|}{12 minutes}     & OLID2019 \\
ALBERT         & \multicolumn{1}{|c|}{70\%} & \multicolumn{1}{|c|}{68.4\%} & \multicolumn{1}{|c|}{0.69} & \multicolumn{1}{c|}{0.68}        &     \multicolumn{1}{c|}{12 minutes}      & OLID2019\\
BERT         & \multicolumn{1}{|c|}{75.6\%} & \multicolumn{1}{|c|}{69.7\%} & \multicolumn{1}{|c|}{0.674} & \multicolumn{1}{c|}{0.64}        &     \multicolumn{1}{c|}{12 minutes}      & OLID2019\\
\textbf{BlazingText}       & \multicolumn{1}{|c|}{71\%}     &  \multicolumn{1}{|c|}{68.34\%} & \multicolumn{1}{|c|}{0.643} & \multicolumn{1}{c|}{0.629}   &  \multicolumn{1}{c|}{40 seconds} & OLID2019 + OLID2020\\

\hline
All UNT           &   \multicolumn{1}{|c|}{42\%}    & \multicolumn{1}{|c|}{40.3\%} & \multicolumn{1}{|c|}{0.30}  &  \multicolumn{1}{c|}{0.287}    &    \multicolumn{1}{c|}{NA}   &  \\
All TIN           &   \multicolumn{1}{|c|}{58\%}    & \multicolumn{1}{|c|}{59.7\%} & \multicolumn{1}{|c|}{0.12}   &  \multicolumn{1}{c|}{0.374}   &   \multicolumn{1}{c|}{NA}    &   \\
\hline
\end{tabular}
\caption{Experimental results of sub-task B (CV = cross-validation; gold = gold test set) on the 2020 test data set}.
\label{tab:taskAResults}
\end{table}

\begin{table}[H]
\centering
\begin{tabular}{llllll|l}
\multirow{2}{*}{\textbf{Model}} & \multicolumn{2}{|c|}{\textbf{Accuracy}}                                & \multicolumn{2}{c|}{F1}                    &        \multirow{2}{*}{\textbf{Runtime}}     &     \multirow{2}{*}{\textbf{Training Data}}  \\
\cline{2-5}
                  & \multicolumn{1}{|c|}{CV}      & \multicolumn{1}{c|}{Gold}    & \multicolumn{1}{c|}{CV}      & \multicolumn{1}{c|}{Gold}      &            \\
\hline
\textbf{RoBERTa}         & \multicolumn{1}{|c|}{88.7\%} & \multicolumn{1}{|c|}{86.6\%} & \multicolumn{1}{|c|}{0.79} & \multicolumn{1}{c|}{0.738}        &     \multicolumn{1}{c|}{12 minutes}  &    OLID2019\\
XLM-RoBERTa         & \multicolumn{1}{|c|}{85\%} & \multicolumn{1}{|c|}{83.3\%} & \multicolumn{1}{|c|}{0.72} & \multicolumn{1}{c|}{0.70}        &     \multicolumn{1}{c|}{12 minutes}     & OLID2019 \\
BERT         & \multicolumn{1}{|c|}{85.3\%} & \multicolumn{1}{|c|}{83.7\%} & \multicolumn{1}{|c|}{0.729} & \multicolumn{1}{c|}{0.67}        &     \multicolumn{1}{c|}{12 minutes}      & OLID2019\\
ALBERT         & \multicolumn{1}{|c|}{81\%} & \multicolumn{1}{|c|}{80.4\%} & \multicolumn{1}{|c|}{0.67} & \multicolumn{1}{c|}{0.657}        &     \multicolumn{1}{c|}{12 minutes}      & OLID2019\\
\textbf{BlazingText}       & \multicolumn{1}{|c|}{92\%}     &  \multicolumn{1}{|c|}{90\%} & \multicolumn{1}{|c|}{0.63} & \multicolumn{1}{c|}{0.573}   &  \multicolumn{1}{c|}{40 seconds} & OLID2019 + OLID2020\\

\hline
All UNT           &   \multicolumn{1}{|c|}{12\%}    & \multicolumn{1}{|c|}{11.75\%} & \multicolumn{1}{|c|}{0.106}  &  \multicolumn{1}{c|}{0.101}    &    \multicolumn{1}{c|}{NA}   &  \\
All TIN           &   \multicolumn{1}{|c|}{88\%}    & \multicolumn{1}{|c|}{88.75\%} & \multicolumn{1}{|c|}{0.468}   &  \multicolumn{1}{c|}{0.47}   &   \multicolumn{1}{c|}{NA}    &   \\
\hline
\end{tabular}
\caption{Experimental results of sub-task B (CV = cross-validation; gold = gold test set) on the 2019 test data set}.
\label{tab:taskAResults}
\end{table}



\section{Results Analysis}

In this section, We have performed an error analysis to identify the shortcomings of the systems we have deployed that achieved the best results, in particular, the BERT based models. The miss-classified instance on both test datasets were investigated and analyzed to identify the challenges and highlight insights regarding the performance of the models.

The results shows that RoBERTa model was top performing model by a significant margin for both Task A and Task B on both years 2019 and 2020 Offenseval test datasets.  

The models seems to be more sensitive to the common words especially the profanity words appeared in offensive tweets such that many misclassified non-offesnive tweets contains some kind of these words are detected as offensive tweets. Our observation however is that some of classification of the test tweets are arguable and might entail negative connotation based on the medium, the context they appear on and the definition of what offensive language entails.  For example the tweet example \textit{"Oh my god drop them that's digusting"} is annotated as a non-offensive tweet which is reasonable if it appear between a teenager or friends community network but also can be offensive in the context of professional or work community. 

The best performing model shortcomings were mainly under the following tweet categories:
\begin{enumerate}
\item Sarcastic tweet where the model showed both false positive and false negative examples for tweets where the user expressing an opinion by stating an idea while meaning another.
\item  Words that was mentioned only in non offensive context but appeared in an offensive context during test. For example, the word (nauseous) is not recognized as offensive since it only exists in a positive tweet in the fine-tuning training data and it is not part of the BERT model vocabulary.

\end{enumerate}

In table \ref{tab:misclassifiedex} shows some misslabeled examples that follows under the aforementioned categories. 

\begin{table}[ht]
    \centering
    \begin{tabular}{p{0.5\linewidth} | p{0.2\linewidth} | p{0.2\linewidth}}
    \hline
      Tweet                  & \multicolumn{1}{|c|}{True Label}            & \multicolumn{1}{|c}{Predicted Label}  \\ \hline
      Conservatives just like old \#Garbage bags blowing in the wind, here's an old garbage bag now... URL & 
      \multicolumn{1}{|c|}{1} & \multicolumn{1}{|c}{0} \\ 
      Conservatives We mus pray for liberals. They trooly kno knot watt they do. I was brot up socialist democrat, by my wonderful, but, ignorant parents. I served inda military. I learnd I can serve Christ, too. Eye reel -eyesed CONSERVATIVISM was more closely aligned w God. &
      \multicolumn{1}{|c|}{1} & \multicolumn{1}{|c}{0} \\ 
      He is an alien from the future &
      \multicolumn{1}{|c|}{1} & \multicolumn{1}{|c}{0} \\ \hline
      The hearburn is disgusting &
      \multicolumn{1}{|c|}{0} & \multicolumn{1}{|c}{1} \\ 
      Sooo who gon be my whole heart? cause being single sucks &
      \multicolumn{1}{|c|}{0} & \multicolumn{1}{|c}{1} \\
      @USER Dont let politics ruin your family...its just a show full of liars on both sides (certain side maybe be worse but both suck) &
      \multicolumn{1}{|c|}{0} & \multicolumn{1}{|c}{1} \\ \hline
      
    \end{tabular}
    \caption{Common examples of misclassified Tweets by the models}
    \label{tab:misclassifiedex}
\end{table}

\section{Challenges and Future Work}
Offensive language detection systems are very beneficial for filtering harmful contents on the internet however building machine learning models to tackle the problem is highly constraint by the lack of large amounts of high quality data and missing contextual information that would otherwise completely alter the intention and the classification results. Working on a dataset that also maps the domain and context of the tweet would be an important direction to build a robust model for offensive language detection. Moreover, the increased availability of multimodal datasets recently opens up an opportunity to incorporate other information form images, videos and audio clips for model training which has the potential of overcoming many challenges of building unimodal models.

Recently, researchers at Google announced the SMITH model \cite{yang2020beyond} which outperformed the BERT model. SMITH is developer to capture the semantic relationship between sentences in long documents. We are planning to explore this model and its implications on the problem of offensive language detection. 

While the the ultimate goal of machine learning competition are to optimize for accuracy, developing a model for a production environment is highly attributed to other other factors such as costs. We plan to do further investigation for lightweight variants of BERT and how they compare to BlazingText to get the most optimal speed, accuracy, and size balance that is highly desired for production systems which are mostly under budgeting constraints.  

One major issue we have encountered while working on offensive language datasets is the lack of consensus of what makes a text deemed to be offensive or classifying it under particular offense category. Therefore, we are planning to investigate the issue further for an opportunity to the quality of manual annotations and ultimately improved datasets.
 
\section{Conclusion}
Offensive language detection has become an important and challenging problem to tackle with the widespread use and deep integration of social media of our day to day life. Throughtout this work, we aimed to tackle this challenge with the help of state of the art techniques and applying the necessary proprocessing pipeline for irregular social media texts. We have also an accepted paper in SemEval where we have participated in the OffensEval tasks for multilingual offensive language detection and achieved good results which we also improved during this dissertation. 

Pretrained BERT proved to be a very strong candidate to solve the problem of offensive language detection. By just fine-tuning BERT with an accurate and relatively small dataset, it was able to achieve competitive performance compare to conventional machine learning techniques. Running a machine learning for offensive language detection requires also favors a fast model that can produce accurate and instant inferences on a large scale environments, the BlazingText shows a great performance that was marginally worst the the BERT counterparts but much faster both in training the model as well as when deploying it to make inferences. Moreover, we have tested the multilingual XLM-RoBERTa for the task which achieved comparable results to the uni-lingual model with the advantage of being applicable to many other language. However, multilingual model performance didn't surpass the uni-lingual model performance that we have tested. 

We have also investigated the performance of the model in terms of the running time to train and make inferences. For that an efficient implementation of the FastText called BlazingText that runs on Amazon AWS Sagemaker where used. The results were very promising while gaining a major leap in the time required to train the model, only insignificant loss of accuracy compared to the top performing RoBERTa model.


\newpage

\backmatter

\printbibliography


\chapter{Eidesstattliche Erkl\"arung}

	Hiermit versichere ich, dass ich diese \thesisType{} selbstst\"andig und ohne Benutzung anderer als der angegebenen Quellen und Hilfsmittel angefertigt habe und alle Ausf\"uhrungen, die w\"ortlich oder sinngem\"a\ss{} übernommen wurden, als solche gekennzeichnet sind, sowie, dass ich die \thesisType ~in gleicher oder \"ahnlicher Form noch keiner anderen Pr\"ufungsbeh\"orde vorgelegt habe.

	\vspace{3cm}

	Passau, \thedate

	\vspace{2cm}

	\parbox{8cm}{
		\hrule \strut \theauthor
	}

\end{document}